\documentclass{article} % For LaTeX2e
\usepackage{iclr2026_conference,times}

\usepackage{amsmath,amsfonts,bm}

\def\eqref#1{equation~\ref{#1}}
\def\1{\bm{1}}

\DeclareMathAlphabet{\mathsfit}{\encodingdefault}{\sfdefault}{m}{sl}
\SetMathAlphabet{\mathsfit}{bold}{\encodingdefault}{\sfdefault}{bx}{n}

\usepackage{hyperref}
\usepackage{url}
\usepackage{graphicx}
\usepackage{booktabs}
\usepackage{subcaption}
\usepackage{placeins}
\usepackage{tocloft}
\usepackage{amssymb}

\usepackage{xcolor}

\title{Property-Guided Molecular Generation and Optimization via Latent Flows}

\author{Alexander Arjun Lobo\thanks{Equal Contribution}\hspace{0.4em},    Urvi Awasthi\footnotemark[1] \hspace{0.2em}, \& Leonid Zhukov\\
BCG X AI Science Institute\\
\texttt{\{lobo.alexander, awasthi.urvi, zhukov.leonid\}@bcg.com} \\
}

\iclrfinalcopy % Uncomment for camera-ready version, but NOT for submission.
\begin{document}

\maketitle

% \begin{abstract}
% Despite recent advances in generative models for molecular and materials discovery, practical progress toward useful and targeted candidates remains governed by inverse design problems that are often either computationally intractable or prone to producing incoherent molecular candidates.
% We introduce a modular framework for molecular generation and optimization that operates in a property-shaped latent space.
% This framework combines variational autoencoders to learn continuous molecular representations, flow-based latent generative models to capture the distribution of valid candidates, and differentiable surrogate predictors that enable gradient-based guidance.
% We evaluate the framework on the task of ionic liquid discovery for CO$_2$ capture, a setting characterized by strong trade-offs between solubility, viscosity.
% Across a range of experiments, we find that guided latent flows consistently advance Pareto fronts beyond those obtained through brute-force enumeration, unconditional generation, and latent-space optimization baselines, while maintaining validity and diversity.
% These results suggest that latent flow-based optimization provides a coherent and practical approach for multi-objective molecular design.
% \end{abstract}

\begin{abstract}
Molecular discovery is increasingly framed as an inverse design problem: identifying molecular structures that satisfy desired property profiles under feasibility constraints.
While recent generative models provide continuous latent representations of chemical space, targeted optimization within these representations often leads to degraded validity, loss of structural fidelity, or unstable behavior.
We introduce \textbf{MoltenFlow}, a modular framework that combines property-organized latent representations with flow-matching generative priors and gradient-based guidance.
This formulation supports both conditioned generation and local optimization within a single latent-space framework.
We show that guided latent flows enable efficient multi-objective molecular optimization under fixed oracle budgets with controllable trade-offs, while a learned flow prior improves unconditional generation quality.
\end{abstract}

\section{Introduction}
\label{sec:introduction}

Molecular and materials discovery problems are increasingly framed as inverse design tasks: rather than evaluating a fixed set of candidates, the goal is to identify molecular structures that satisfy specified property profiles under physical and feasibility constraints.
This formulation arises naturally across applications such as drug design, solvent discovery, and functional materials engineering, where target properties are expensive to evaluate and often exhibit non-trivial trade-offs.
A common approach combines large-scale enumeration with learned property predictors to enable rapid virtual screening.
However, such screening-based pipelines scale poorly with chemical space size and offer limited support for systematic refinement of promising candidates \citep{gomez2018automatic,sindt2026screening}.

Recent advances in generative modeling have motivated a shift toward learning continuous representations of chemical space that can be directly sampled or optimized.
Variational autoencoders (VAEs) and related latent variable models provide smooth latent spaces in which discrete molecular structures are embedded, enabling interpolation and gradient-based traversal \citep{gomez2018automatic,jin2018junction,kusner2017grammar}.
Despite their success in unconditional generation, these representations are typically learned to optimize reconstruction fidelity and distributional regularization, rather than alignment with task-specific objectives.
As a result, naïve latent-space optimization often leads to invalid decodings, unstable property trajectories, or collapsed diversity when strong gradients are applied \citep{tripp2020bayesian,eckmann2022limolatentinceptionismtargeted}.

To address these limitations, a broad class of goal-directed molecular design methods has emerged.
Reinforcement learning approaches formulate molecular generation as a sequential decision process \citep{olivecrona2017molecular,zhou2019optimization}, while Bayesian optimization techniques operate over learned latent spaces \citep{griffiths2020constrained,tripp2020bayesian}.
Although effective in specific settings, these methods often suffer from sample inefficiency, sensitivity to reward shaping, or difficulties maintaining diversity under aggressive optimization.
More recently, diffusion-based generative models have demonstrated strong performance in molecular generation \citep{hoogeboom2022equivariant,xu2022geodiff}, and classifier-guided or conditional variants offer principled mechanisms for steering samples toward desired objectives.
However, applying such guidance directly in high-dimensional molecular spaces remains computationally intensive and challenging to control \citep{lin2025tfgflowtrainingfreeguidancemultimodal, chung2023dps}.

In parallel, flow-based and diffusion-based generative models have been proposed as alternatives for learning smooth generative dynamics over molecular spaces \cite{lipman2022flow,albergo2023stochastic,hoogeboom2022equivariant,xu2022geodiff}.
Recent frameworks such as ChemFlow and PropMolFlow learn energy functions or vector fields that directly condition the generative dynamics on target properties \cite{wei2024chemflow,zeng2025propmolflow}, while methods such as MolGuidance introduce guidance terms during sampling in data space \cite{jin2025molguidance}.
Other approaches, including LIMO and related latent-gradient methods, optimize properties by performing gradient-based updates in a fixed latent space, without an explicit generative prior \cite{eckmann2022limolatentinceptionismtargeted}.
Complementary lines of work in latent-space Bayesian optimization, such as LOL-BO and NF-BO, achieve sample efficiency through probabilistic surrogate modeling but require fitting additional models over high-dimensional latent spaces \cite{maus2023lolbo,lee2025nfbo}.

In this work, we introduce \textbf{MoltenFlow}, a framework that occupies a distinct point in this design space.
Rather than conditioning generative dynamics directly or relying on a second surrogate over latent space, MoltenFlow combines representation-level property organization with a learned latent flow prior to support controllable inverse molecular design.
This design targets the regime where optimization must be both sample-efficient and stable, enabling systematic advancement of multi-objective trade-offs while remaining close to the learned data manifold. Across multi-objective molecular optimization tasks, we show that guided latent flows advance Pareto fronts while exposing an explicit trade-off between objective improvement and structural fidelity.
\section{Methods}
\label{sec:method}

\begin{figure*}[t]
    \centering
    \includegraphics[width=0.95\linewidth]{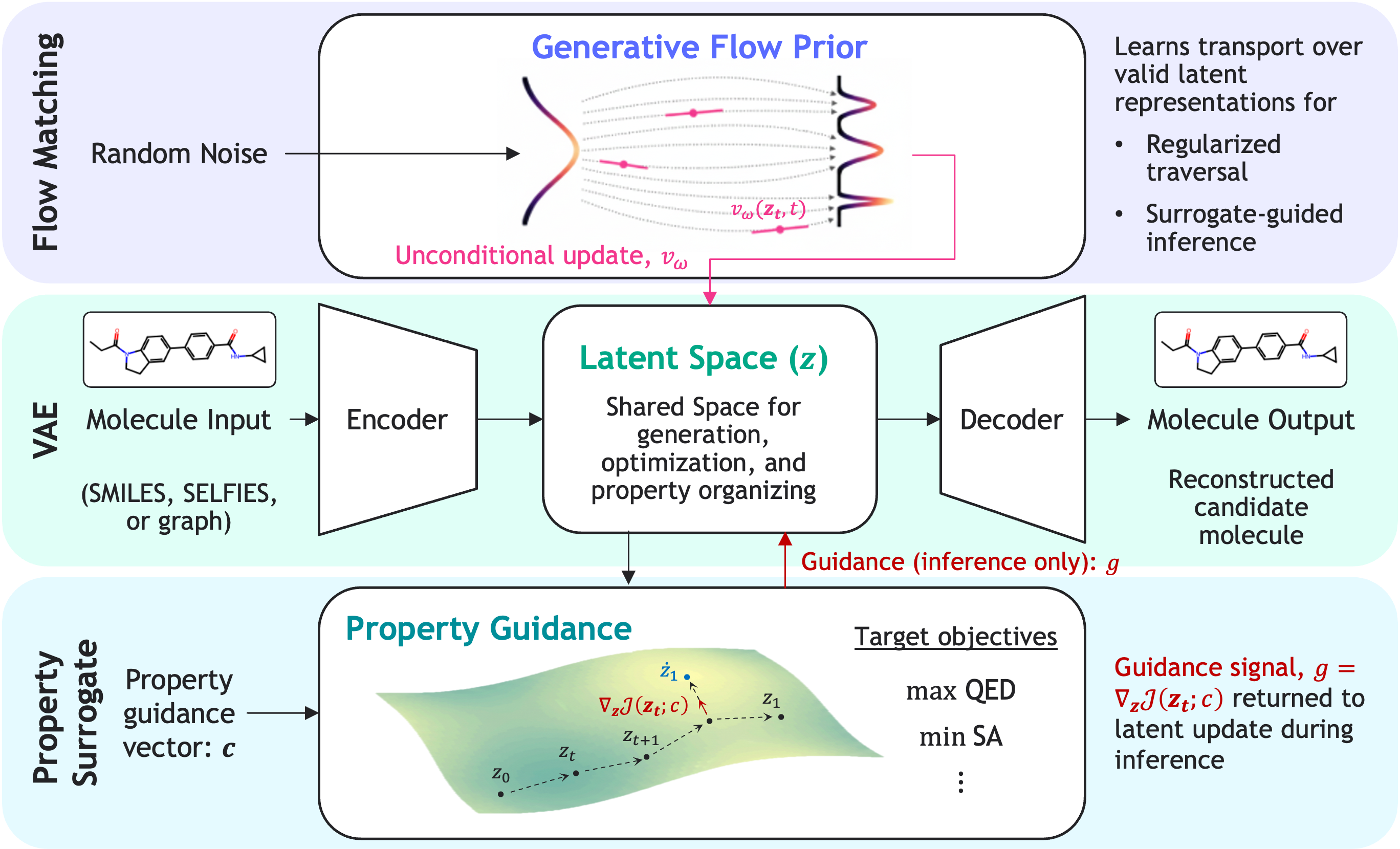}
    \caption{\textbf{MoltenFlow overview.}
A VAE maps discrete molecular inputs \(x\) to continuous latent representations \(z\) and reconstructs molecules via a decoder.
The latent space is \emph{organized} through auxiliary property prediction: a differentiable property surrogate \(f_\psi(z)\) induces an objective \(\mathcal{J}(z; c)\) (e.g., target values or directional goals specified by a guidance vector \(c\)), and provides a guidance signal \(g=\nabla_z \mathcal{J}(z_t; c)\) during inference.
In parallel, a flow-matching model learns a time-dependent vector field $v_\omega(z,t)$ that transports samples from a simple base distribution (random noise) to the empirical distribution of valid latent representations. \emph{The flow-based latent transport visualization is adapted from ~\citet{sabour2025alignflowscalingcontinuoustime}, and the property-guidance schematic is inspired by ChemFlow~\citep{wei2024chemflow}}.
}
    \label{fig:moltenflow-figure}
\end{figure*}

We present \textbf{MoltenFlow}, a modular framework for molecular generation and optimization in continuous latent space. We build on latent variable generative modeling, where a VAE is used to learn a continuous representation that captures the manifold of valid chemical compounds, such that nearby latent representations correspond to structurally similar molecules. To make representations more suitable for inverse design, we further organize the latent space through auxiliary property prediction, allowing gradients from property supervision to reshape the encoder. In parallel, we learn a generative prior over latent space using flow matching, which models the distribution of valid latent representations and regularizes traversal to remain near the data manifold.

This combination produces a latent space and traversal method that is both property-aware and generative, enabling two complementary inference modes: \textit{conditioned generation} and \textit{optimization}. In conditioned generation mode, molecular candidates are produced by transforming noise into valid latent representations guided by desired property objectives. In optimizer mode, existing molecules are encoded, perturbed with noise, and then guided toward improved latent representations before decoding. In both cases, property objectives may be specified as explicit targets or as directional objectives, such as maximizing or minimizing a property of interest. Compared to prior latent-space optimization and reinforcement learning approaches, this formulation yields stable continuous-time guidance dynamics with an explicit guidance parameter that trades off objective improvement against remaining near the learned latent data manifold, without requiring sequential decision-making or task-specific policy training.

\subsection{Latent Representation Learning}

Let \( x \in \mathcal{X} \) denote a molecular structure represented in a discrete format (e.g., SMILES or SELFIES; see Appendix~\ref{app:mol-representations} for representation details).
We learn a continuous latent representation \( z \) using a variational autoencoder (VAE). Here, \( z \) is a set of token-level latent representations \( z \in \mathbb{R}^{K \times d} \) produced by attention-based pooling over encoder outputs. When a vector-valued latent representation is required, we denote the pooled representation as \( \bar z \in \mathbb{R}^{d} \).
The generative model is defined by a prior \( p(z) \) and a decoder \( p_\theta(x \mid z) \), while inference is performed using a variational posterior \( q_\phi(z \mid x) \), where $\theta$ and $\phi$ parameterize the decoder and encoder, respectively.
Model parameters are learned by maximizing the evidence lower bound (ELBO):
\begin{equation}
\label{eq:vae-objective}
\mathcal{L}_{\mathrm{VAE}}(x)
=
\mathbb{E}_{q_\phi(z \mid x)}\!\left[\log p_\theta(x \mid z)\right]
-
\mathrm{KL}\!\left(q_\phi(z \mid x)\,\|\,p(z)\right)
\end{equation}

We parameterize $q_\phi(z\mid x)$ as a diagonal Gaussian and use the reparameterization trick,
$z=\mu_\phi(x)+\sigma_\phi(x)\odot\epsilon$, $\epsilon\sim\mathcal{N}(0,I)$, to optimize this objective by backpropagation.

This objective encourages latent representations that support accurate reconstruction while remaining close to the prior distribution.
However, this representation is optimized primarily for reconstruction fidelity and distributional regularization; it does not explicitly encode task-relevant semantics.

We denote the induced latent data distribution (aggregated posterior) by
$q_\phi(z)=\mathbb{E}_{x\sim p_{\mathrm{data}}}\,q_\phi(z\mid x)$.
In later stages, we treat samples $z_1\sim q_{\phi'}(z\mid x)$ from the training set as samples from the target latent distribution $p_1(z)=q_{\phi'}(z)$, where $\phi'$ denotes the encoder parameters after property-oriented fine-tuning (Section~\ref{sec:prop-oriented}).

\subsection{Property-Oriented Latent Space Transformation}
\label{sec:prop-oriented}

To make latent space traversal meaningful for inverse design, we augment latent representation learning with property supervision.
Let \( y \in \mathbb{R}^P \) denote a vector of molecular properties of interest, and let \( f_\psi : \mathbb{R}^{K \times d} \rightarrow \mathbb{R}^P \) be a differentiable surrogate model that predicts properties from latent representations.

We introduce an auxiliary property prediction loss
\begin{equation}
\label{eq:finetune-loss}
\mathcal{L}_{\mathrm{prop}}(z, y) = \ell\!\left(f_\psi(z),\, y\right)
\end{equation}
where \( \ell(\cdot,\cdot) \) is an appropriate regression loss.
During training, this loss is combined with the VAE objective to yield
\begin{equation}
\mathcal{L}(x, y)
=
\mathcal{L}_{\mathrm{VAE}}(x)
+
\lambda\, \mathcal{L}_{\mathrm{prop}}(z, y)
\end{equation}
with \( z \sim q_\phi(z \mid x) \) and \( \lambda \geq 0 \) controlling the strength of property supervision.

Allowing gradients of \( \mathcal{L}_{\mathrm{prop}} \) to backpropagate through the encoder \( q_\phi \) reshapes the latent geometry such that local directions in latent space correlate with changes in predicted properties \cite{gomez2018automatic}.
This process can be interpreted as transforming the latent representation into one that is better aligned with downstream optimization objectives.
The resulting latent space remains continuous and smooth, but is additionally structured by property-relevant variation.

\subsection{Latent Flow Matching}

While property-oriented latent representations improve local interpretability, they do not guarantee that latent trajectories remain within realistic regions of chemical space. Syntactically robust representations such as SELFIES ensure valid decoding but do not prevent drift into low-density regions; we therefore learn a flow-based prior over latent space to model the empirical distribution and regularize trajectories toward the data manifold.

Flow matching seeks to learn a time-dependent vector field
\begin{equation}
v_\omega : ( \mathbb{R}^{K \times d} \times [0,1]) \rightarrow \mathbb{R}^{K \times d},
\end{equation}
where $\omega$ denotes the parameters of the flow model, which transports samples from a simple base distribution \( p_0(z) \) (e.g., Gaussian noise) to a target distribution \( p_1(z) \) (the empirical distribution of latent codes).
Training proceeds by sampling endpoints $z_0\sim p_0$, $z_1\sim p_1$, and a time $t\sim\mathcal{U}[0,1]$.
For notational convenience, we write $z_t := z(t)$ to denote the latent state at time $t \in [0,1]$. We use the linear probability path
\begin{equation}
z_t = (1-t)z_0 + t z_1,
\qquad
\frac{d z_t}{dt}=z_1-z_0
\end{equation}
Flow matching then fits the time-dependent vector field by regression:
\begin{equation}
\label{eq:flow-loss}
\mathcal{L}_{\mathrm{FM}}(\omega)=
\mathbb{E}_{z_0\sim p_0,\,z_1\sim p_1,\,t\sim \mathcal{U}[0,1]}
\left[\left\|v_\omega(z_t,t)-(z_1-z_0)\right\|_2^2\right]
\end{equation}

The learned vector field captures the geometry of the latent data distribution, providing a manifold-aware generative model.
Sampling from this flow yields latent representations that remain close to regions supported by training data, while avoiding the need for explicit likelihood computation or invertibility constraints.

At inference time, the learned prior samples latents by integrating the ODE
$\dot z(t)=v_\omega(z(t),t)$ from $t=0$ to $1$ with initial condition $z(0)\sim p_0$; the final latent $z(1)$ is decoded to a molecule using the VAE decoder.

\subsection{Guided Latent Dynamics for Generation and Optimization}
\label{sec:guidance}

The final component of MoltenFlow is a guidance mechanism that combines the learned latent flow with gradients derived from surrogate objectives.
Let \( f_\psi(z) \in \mathbb{R}^P \) denote the surrogate-predicted properties associated with a latent representation \( z \), and let \( c \) denote a vector specifying desired targets or objective preferences for these properties.
We define a scalar objective function \( \mathcal{J}(z; c) \), parameterized by \( c \), and compute its gradient with respect to latent space:
\begin{equation}
g(z) = \nabla_z \mathcal{J}(z; c)
\end{equation}

We interpret $\mathcal{J}(z;c)$ as a loss function to be minimized; maximization objectives are handled by defining $\mathcal{J}$ with an appropriate negative sign (see Appendix~\ref{app:objective_construction}).
During inference, we modify the latent flow dynamics by combining the generative velocity with the objective gradient:
\begin{equation}
\label{eq:flow_dynamics}
\dot{z}(t)
=
v_\omega(z(t), t)
-
\gamma\, g(z(t)),
\end{equation}
where \( \gamma > 0 \) controls the strength of guidance.
This formulation yields a continuous-time dynamical system that balances two competing effects: movement along the learned data manifold and movement toward improved objective values specified by \( c \). Using a learned latent prior in Eq.~\ref{eq:flow_dynamics} is critical: direct gradient ascent on surrogate objectives often pushes latents into low-density regions where decoding becomes unreliable. The flow term \(v_\omega(z,t)\) regularizes these updates by biasing trajectories toward the support of \(q_{\phi'}(z)\), improving feasibility for a given guidance strength \(\gamma\).

By varying the form of \( \mathcal{J} \), the conditioning vector \( c \), and the initial condition \( z(0) \), the same guided dynamics support multiple inference modes.
Starting from samples drawn from the base distribution yields generation directed toward specified property objectives, while initializing from the latent encoding of an existing molecule enables local optimization.
In both cases, the learned latent flow regularizes traversal, ensuring that optimization proceeds through regions of latent space associated with valid molecular structures.
%%%%%%%%%%%%%%%%%%%%%%%%%%%%%%%%%%%%%%%%
% Experiments
%%%%%%%%%%%%%%%%%%%%%%%%%%%%%%%%%%%%%%%%
\section{Experiments}
\label{sec:experiments}

% \todo[inline]{@Alex: Make sure that preface for experiments matches content. Also, reduce the sections in experiments. A good idea is to frame the experiments as molecular optimization and molecular generation. First will share the results of the hpo, budgetted optimization, and global pareto front improvement. Molecular generation will include Urvi's ablations and (if time) and analysis of conditioned vs unconditioned generation}

% This section evaluates MoltenFlow across a sequence of optimization settings designed to test its ability to (i) generate valid molecular structures, (ii) perform stable, property-aware optimization, and (iii) advance multi-objective trade-offs while remaining close to the data manifold.
% We begin with a large-scale synthetic benchmark that enables statistically robust Pareto analysis, followed by experiments on experimentally motivated property datasets, and conclude with an optional application to ionic liquid design.
% Throughout, we emphasize optimizer-mode evaluation, where existing molecules are iteratively improved under competing objectives.

%%%%%%%%%%%%%%%%%%%%%%%%%%%%%%%%%%%%%%%%
% `Optimization Mode` section
%%%%%%%%%%%%%%%%%%%%%%%%%%%%%%%%%%%%%%%%

\subsection{Multi-objective Molecular Optimization}

We evaluate MoltenFlow’s optimizer mode through two complementary studies that assess performance in simulated practical discovery workflows and advancing global Pareto fronts.
Unless otherwise noted, all optimization experiments use a fixed MoltenFlow configuration selected via Bayesian hyperparameter optimization, operate on the same learned latent representation with a shared surrogate model, and employ SELFIES to ensure syntactic validity and isolate optimization dynamics.
Details of the search space and tuning procedure are provided in Appendix~\ref{app:budgeted-hpo}.

%%%%%%%%%%%%%%%%%%%%%%%%%%%%%%%%%%%%%%%%
% Budgeted Optimization
%%%%%%%%%%%%%%%%%%%%%%%%%%%%%%%%%%%%%%%%

\subsubsection{Budgeted optimization}

We first evaluate MoltenFlow in a budgeted optimization setting that mirrors practical molecular discovery workflows, where candidate evaluations are expensive and optimization must proceed under a fixed oracle budget.
In this regime, MoltenFlow consistently achieves larger and faster Pareto-front improvements across seeds than both latent-space Bayesian optimization and unregularized gradient-based optimization, while maintaining perfect validity.
This advantage emerges early in the optimization process and persists across random initializations, highlighting the utility of combining surrogate gradients with a learned, manifold-aware latent flow.

Each optimization run begins from a small set of randomly selected molecules and proceeds sequentially, proposing new candidates until the oracle budget is exhausted (see Appendix~\ref{app:budgeted-details} for details).
We compare MoltenFlow against two latent-space Bayesian optimization baselines: one with a multi-output Gaussian process and the other with independent Gaussian processes per objective, both with qEHVI acquisition. We also compare against a gradient ascent ablation that removes the flow prior and follows surrogate gradients directly. All methods are evaluated under the same fixed oracle budget; MoltenFlow leverages a pretrained surrogate (offline setting), while Bayesian optimization fits its surrogate online.

% \begin{figure}[t]
% \centering
% \includegraphics[width=0.92\linewidth]{figures/hvi_convergence_random_B100.pdf}
% \caption{\textbf{Budgeted optimization performance.}
% Hypervolume improvement (HVI) as a function of oracle calls under a fixed budget.
% Curves show mean performance across random seeds, with 90\% bootstrap confidence intervals.
% }
% \label{fig:budgeted_hvi}
% \end{figure}

% \begin{figure}[t]
% \centering
% \includegraphics[width=0.88\linewidth]{figures/pareto_density_random.pdf}
% \caption{\textbf{Density structure of Pareto-optimal solutions in objective space after budgeted optimization.}
% MoltenFlow produces dense and stable fronts concentrated in the high-QED, low-SA region, whereas Bayesian optimization baselines yield sparser and more variable solution sets across runs.
% }
% \label{fig:pareto_density}
% \end{figure}

\begin{figure}[t]
\centering

\begin{minipage}[t]{0.51\linewidth}
    \centering
    \includegraphics[width=\linewidth]{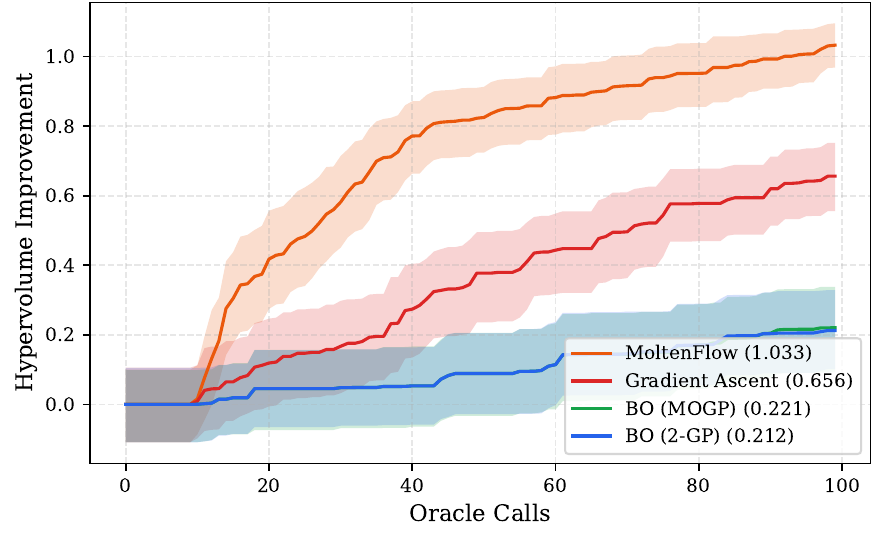}
    \captionof{figure}{
    \textbf{Hypervolume improvement (HVI)} as a function of oracle calls under a fixed budget.
    Curves show mean performance across random seeds, with 90\% bootstrap confidence intervals.}
    \label{fig:budgeted_hvi}
\end{minipage}
\hfill
\begin{minipage}[t]{0.44\linewidth}
    \centering
    \includegraphics[width=\linewidth]{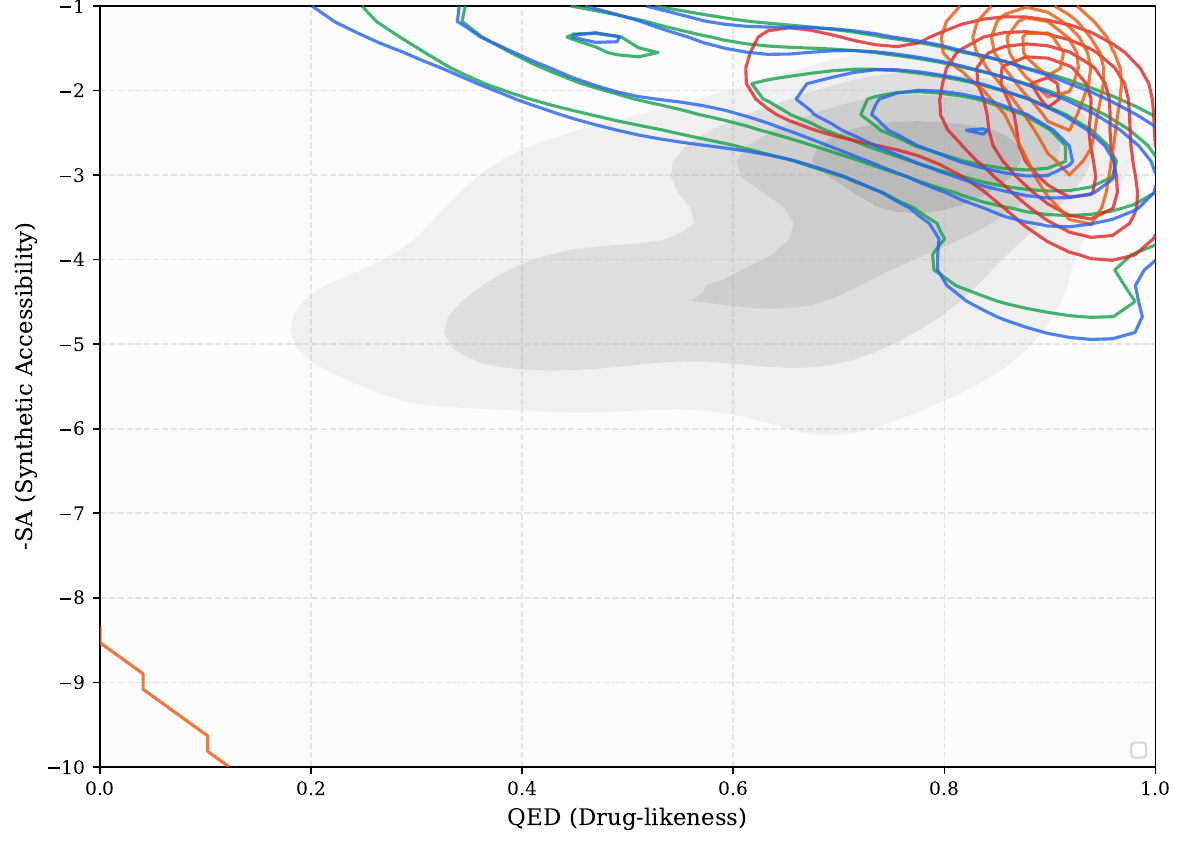}
    \captionof{figure}{
    \textbf{Densities of Pareto fronts} after budgeted optimization.
    MoltenFlow yields dense, stable fronts, while BO baselines yield sparser, more variable fronts.}
    \label{fig:pareto_density}
\end{minipage}

\end{figure}

Figure~\ref{fig:budgeted_hvi} illustrates the evolution of hypervolume improvement as a function of oracle calls.
MoltenFlow exhibits rapid early gains, reaching a substantially improved Pareto front within a small fraction of the available budget.
In contrast, both Bayesian optimization variants improve more slowly and exhibit greater variability across runs, consistent with the challenges of fitting Gaussian process models in moderate-dimensional latent spaces.
The gradient ascent ablation performs competitively at early stages but consistently underperforms MoltenFlow, indicating that flow-based regularization contributes meaningfully beyond the surrogate gradient direction alone.

The qualitative structure of the optimized solution sets further distinguishes the methods.
As shown in Figure~\ref{fig:pareto_density}, MoltenFlow concentrates probability mass in the high-QED, low-SA region of objective space, producing dense and stable Pareto fronts across random seeds.
Bayesian optimization methods, by contrast, yield sparser fronts with higher variance, reflecting sensitivity to acquisition noise and surrogate uncertainty under limited budgets.

From a computational perspective, MoltenFlow is substantially more efficient than Bayesian optimization (see Appendix~\ref{app:budget-extended-results}).
By directly leveraging gradients from a differentiable surrogate and avoiding repeated Gaussian process fitting and acquisition maximization, MoltenFlow achieves comparable or better improvements with significantly lower wall-clock time per run.
This efficiency advantage becomes increasingly relevant as oracle budgets grow or optimization is embedded within iterative discovery pipelines.

While the budgeted setting highlights MoltenFlow’s practical efficiency under limited evaluations, we next examine how guided latent flows systematically reshape the Pareto front itself and the trade-offs induced by varying the guidance strength.

%%%%%%%%%%%%%%%%%%%%%%%%%%%%%%%%%%%%%%%%
% Pareto front advancement
%%%%%%%%%%%%%%%%%%%%%%%%%%%%%%%%%%%%%%%%

\subsubsection{Pareto front advancement}
\label{sec:pareto-advancement}

We study MoltenFlow’s ability to advance the Pareto front for drug-like molecules by jointly optimizing Quantitative Estimate of Drug-likeness (QED; higher is better) and Synthetic Accessibility (SA; lower is better) on ZINC250K.
Guided latent flows induce a clear and interpretable trade-off regime: for a well-chosen range of guidance strengths, MoltenFlow substantially expands the Pareto front while preserving molecular validity and scaffold diversity, whereas overly aggressive guidance leads to distributional drift and structural collapse.

\begin{figure}[t]
\centering
\includegraphics[width=\linewidth]{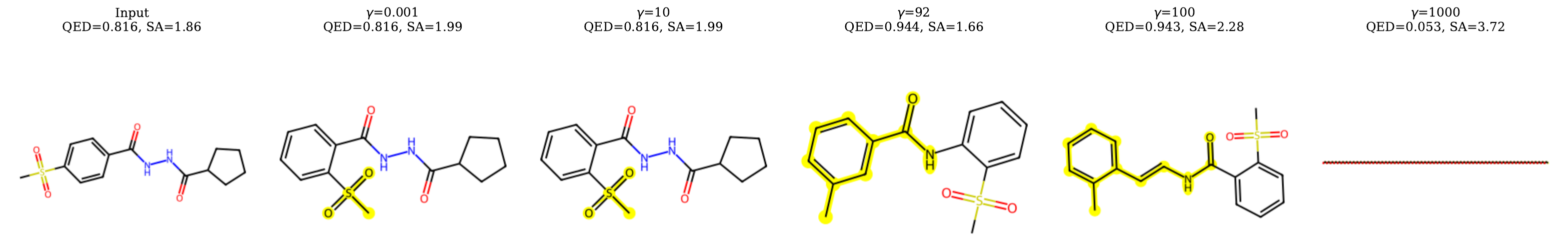}
\caption{\textbf{Qualitative effect of guidance strength on optimization.}
As the guidance strength $\gamma$ increases, MoltenFlow produces progressively larger structural changes.
Small $\gamma$ yields minimal edits dominated by the latent flow prior, intermediate $\gamma$ induces coherent improvements in QED with reasonable SA, and large $\gamma$ leads to aggressive modifications indicative of over-optimization.
}
\label{fig:molecule_progression}
\end{figure}

Figure~\ref{fig:molecule_progression} provides a qualitative view of this behavior.
At very small guidance strengths, optimized molecules remain very similar to their inputs, reflecting dominance of the learned latent flow prior.
As guidance increases, structural changes emerge gradually and coherently, improving QED without immediately sacrificing synthesizability.
In contrast, excessively large guidance produces abrupt edits that compromise structural integrity, consistent with over-optimization beyond the surrogate’s reliable regime.

A differentiable surrogate model predicts QED and SA from latent representations and provides gradients used to guide optimization.
Optimization begins from latent encodings of molecules near the empirical Pareto front, injects modest Gaussian noise to encourage local exploration, and integrates guided dynamics that combine the learned latent flow velocity with normalized property gradients.
The only degree of freedom varied in this study is the guidance strength parameter~$\gamma$, which controls the relative influence of the surrogate objective versus the generative flow prior.

\begin{figure*}[t]
\centering
\includegraphics[width=\linewidth]{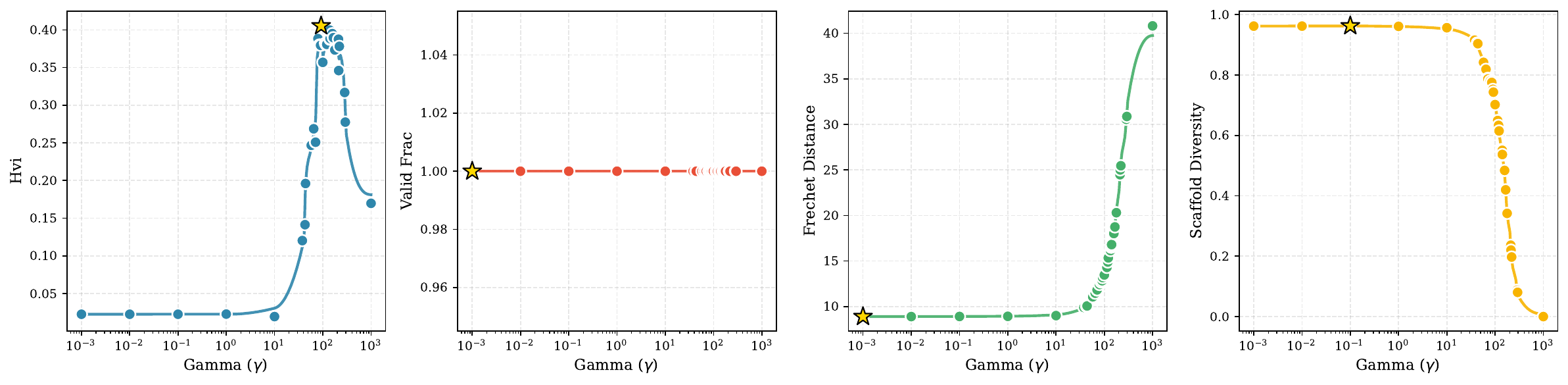}
\caption{\textbf{Effect of guidance strength on optimization and distributional properties.}
Each panel shows a metric as the guidance strength $\gamma$ increases (log scale). From left to right: hypervolume improvement (HVI), validity, Fr\'echet distance (FD-FP), and scaffold diversity.
HVI increases sharply in an intermediate regime, indicating effective Pareto-front advancement, while excessive guidance leads to rising Fr\'echet distance and collapsing diversity, signaling over-optimization.
Starred points denote the $\gamma$ achieving the best value for each metric.
}
\label{fig:hvi_scaffold_tradeoff}
\end{figure*}

\begin{figure}[t]
  \centering

  % -------- Left: Main Pareto figure --------
  \begin{subfigure}[t]{0.65\linewidth}
    \vspace{0pt}
    \centering
    \includegraphics[width=\linewidth]{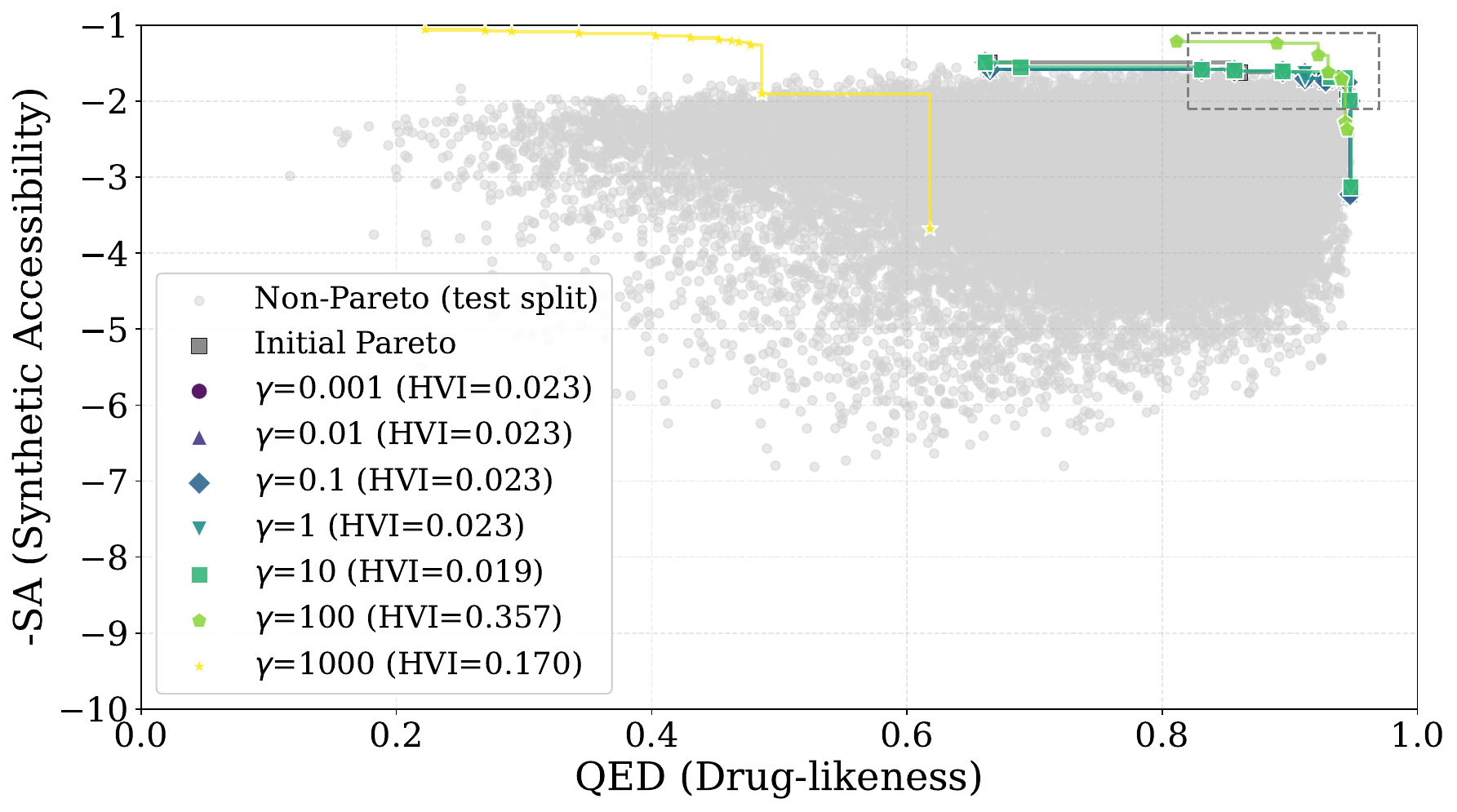}
    \label{fig:pareto_comparison}
  \end{subfigure}
  \hfill
  % -------- Right: Zoomed inset (top) + table (bottom) --------
  \begin{subfigure}[t]{0.33\linewidth}
    \vspace{0pt}
    \centering

    % Zoomed-in view (top)
    \vspace{2pt}
    \includegraphics[width=0.85\linewidth]{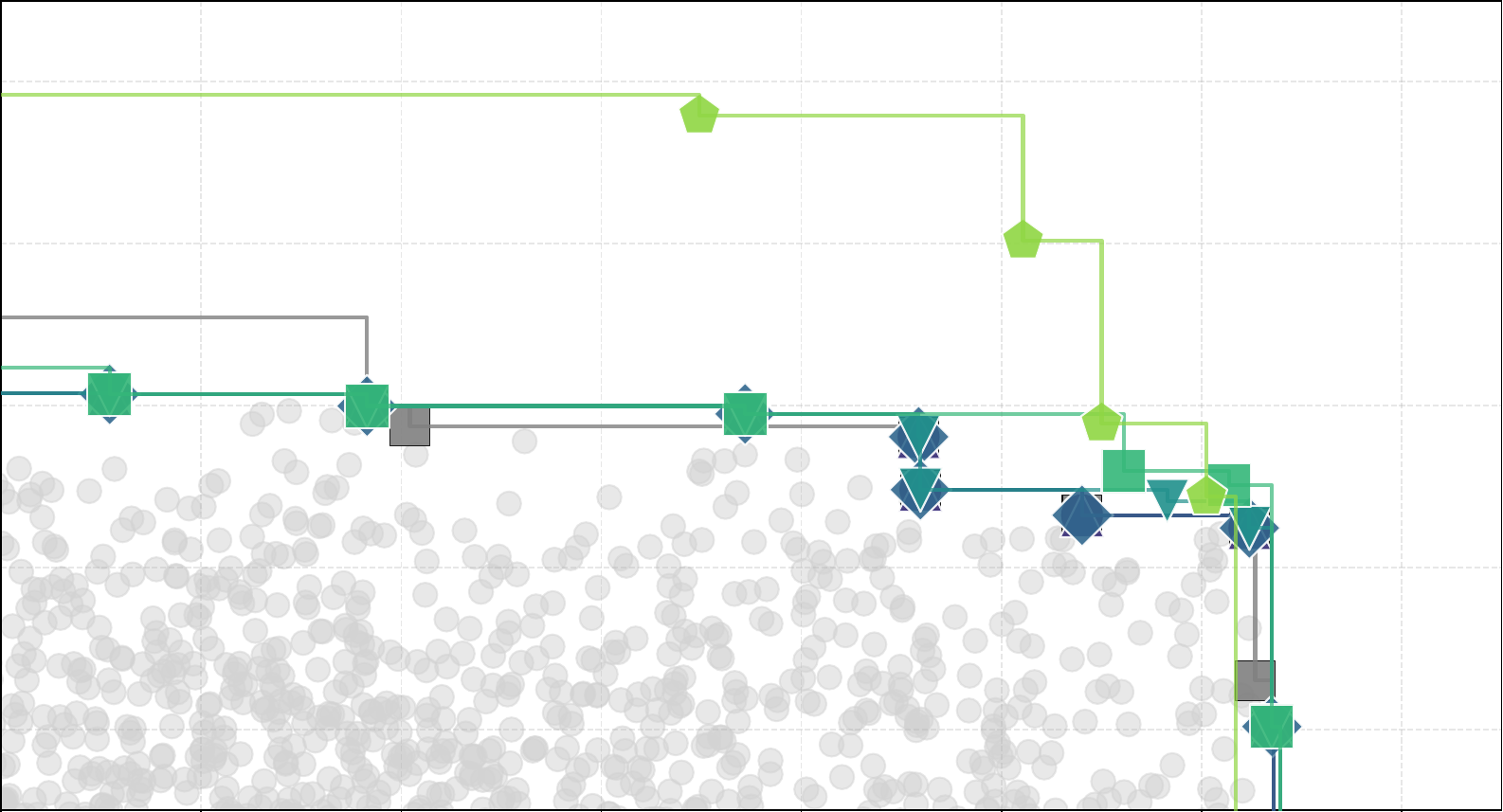}

    \vspace{2pt}

    % Table (bottom)
    \tiny
    \vspace{2pt}

    \setlength{\tabcolsep}{3pt}
    \renewcommand{\arraystretch}{1.05}
    \begin{tabular}{lccc}
      \toprule
      $\gamma$ & HVI (\%)$\uparrow$ & Scaf. Div.$\uparrow$ & FD-FP$\downarrow$ \\
      \midrule
      $0.001$ & $0.39$ & $0.962$ & $8.9$ \\
      $0.1$   & $0.39$ & $0.963$ & $8.9$ \\
      $1$     & $0.39$ & $0.961$ & $8.9$ \\
      $10$    & $0.33$ & $0.956$ & $9.0$ \\
      $\mathbf{100}$ & $\mathbf{6.13}$ & $0.702$ & $13.5$ \\
      $1000$  & $2.91$ & $0.000$ & $40.8$ \\
      \bottomrule
    \end{tabular}

    \label{tab:gamma_sweep}
  \end{subfigure}

  \vspace{4pt}
  \caption{\textbf{Pareto-front advancement and guidance-strength effects.}
  (\textbf{Left}) Optimized Pareto fronts obtained at different guidance strengths $\gamma$.
  At intermediate $\gamma$ (\textcolor{lime}{green}), MoltenFlow produces fronts that strictly dominate the original, expanding into the high-QED, low-SA region while retaining a broad set of trade-offs.
  (\textbf{Right}) Zoomed view of the high-QED/low-SA corner (top) and summary metrics across guidance strengths (bottom), illustrating the trade-off between hypervolume improvement, scaffold diversity, and distributional shift (see Appendix~\ref{app:eval-metrics} for metric definitions).
}
  \label{fig:pareto_and_gamma}
\end{figure}

Across a wide sweep of $\gamma$ values, a consistent pattern emerges (Figure~\ref{fig:hvi_scaffold_tradeoff} and~\ref{fig:pareto_and_gamma}).
At low $\gamma$, trajectories remain close to the data manifold, yielding negligible Pareto expansion but near-perfect scaffold preservation.
As $\gamma$ increases into an intermediate regime, MoltenFlow reliably advances the Pareto front into regions of higher QED and lower SA, achieving the largest hypervolume gains while maintaining substantial scaffold diversity.
Beyond this regime, further increasing $\gamma$ degrades performance: diversity collapses and distributional metrics indicate significant departure from the training manifold.

% \begin{table}[htpb]
% \centering
% \small
% \setlength{\tabcolsep}{4pt}
% \caption{\textbf{Summary of optimization and distributional behavior across guidance strengths.}
% Hypervolume improvement (HVI) quantifies Pareto-front expansion, while scaffold diversity and Fr\'echet distance capture structural diversity and distributional shift, respectively.
% Intermediate guidance strengths yield the largest Pareto gains while retaining substantial diversity; extreme guidance leads to over-optimization and collapse.
% }
% \label{tab:gamma_sweep}
% \begin{tabular}{lccc}
% \toprule
% $\gamma$ & HVI (\%)$\uparrow$ & Scaffold Div.$\uparrow$ & Fr\'echet Dist.$\downarrow$ \\
% \midrule
% $0.001$ & $0.39$ & $0.962$ & $8.9$ \\
% $0.1$ & $0.39$ & $0.963$ & $8.9$ \\
% $1$ & $0.39$ & $0.961$ & $8.9$ \\
% $10$ & $0.33$ & $0.956$ & $9.0$ \\
% $\mathbf{100}$ & $\mathbf{6.13}$ & $0.702$ & $13.5$ \\
% $1000$ & $2.91$ & $0.000$ & $40.8$ \\
% \bottomrule
% \end{tabular}
% \end{table}

Figure~\ref{fig:pareto_and_gamma} visualizes the resulting Pareto fronts.
At the optimal guidance strength, the optimized front strictly dominates the original test-set front, expanding into the high-QED, low-SA region that corresponds to more attractive drug-like candidates.
Importantly, this expansion is not achieved by collapsing to a narrow set of solutions, but by producing a broader front that reflects meaningful trade-offs between objectives.

% \begin{figure}[t]
% \centering
% \includegraphics[width=0.9\linewidth]{figures/pareto_comparison.pdf}
% \caption{\textbf{Pareto-front advancement under guided latent optimization.}
% Grey points denote non-Pareto test molecules, and grey squares denote the initial Pareto front.
% Colored markers show optimized Pareto fronts obtained at different guidance strengths.
% At the tuned intermediate $\gamma$ (green), MoltenFlow produces a front that strictly dominates the original, expanding into the high-QED, low-SA region while retaining a broad set of trade-offs.
% }
% \label{fig:pareto_comparison}
% \end{figure}

% To identify a representative operating point, we employ a two-stage search over $\gamma$: an initial coarse sweep followed by Bayesian optimization to refine around the most promising region.
% This procedure consistently selects an intermediate $\gamma$ value that balances improvement against structural fidelity.
% Further details of the search space and tuning procedure are provided in Appendix~\ref{app:pareto-hpo}.

Taken together, these results demonstrate that MoltenFlow enables controlled and interpretable multi-objective optimization.
The guidance strength parameter provides a continuous knob between conservative, scaffold-preserving refinement and more aggressive exploration, making explicit the trade-offs inherent in latent-space molecular optimization.

%%%%%%%%%%%%%%%%%%%%%%%%%%%%%%%%%%%%%%%%
% Generation Section
%%%%%%%%%%%%%%%%%%%%%%%%%%%%%%%%%%%%%%%%

\subsection{Molecular generation and ablations}
\label{sec:ablations}

We next examine MoltenFlow in a pure generation setting, with surrogate guidance disabled at inference.
These experiments are not intended to demonstrate state-of-the-art unconditional generation performance, but rather to isolate the contribution of two design choices: learning a flow-based latent prior and selecting an appropriate molecular string representation.
The central takeaway is that a learned latent flow prior produces samples that better align with the support of the training distribution than a standard VAE prior, and that this effect is amplified by using a syntactically robust representation such as SELFIES.

\begin{figure*}[t]
\centering
\begin{subfigure}[t]{0.32\linewidth}
  \centering
  \includegraphics[width=\linewidth]{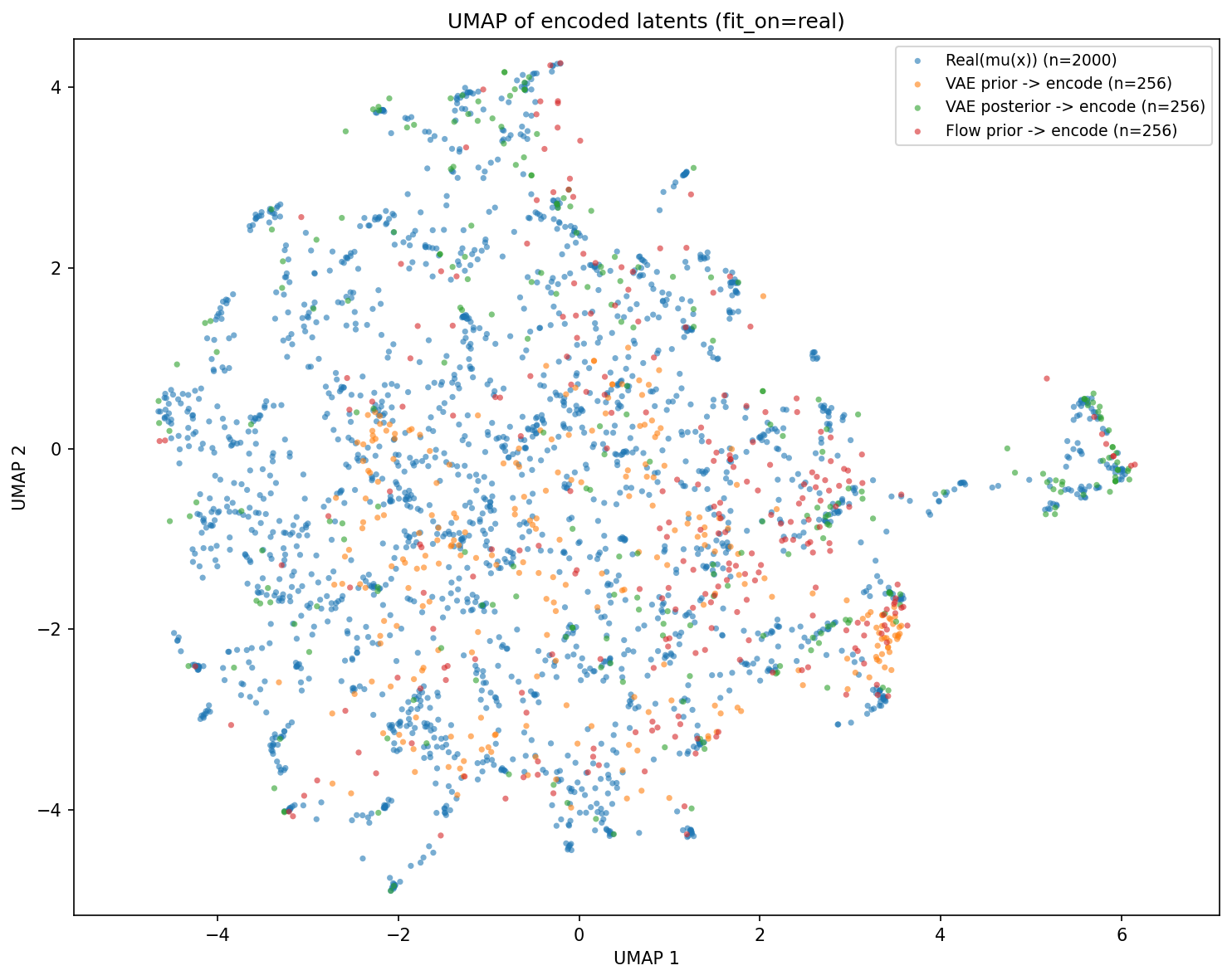}
  \caption{Overlay (fit on real).}
\end{subfigure}\hfill
\begin{subfigure}[t]{0.32\linewidth}
  \centering
  \includegraphics[width=\linewidth]{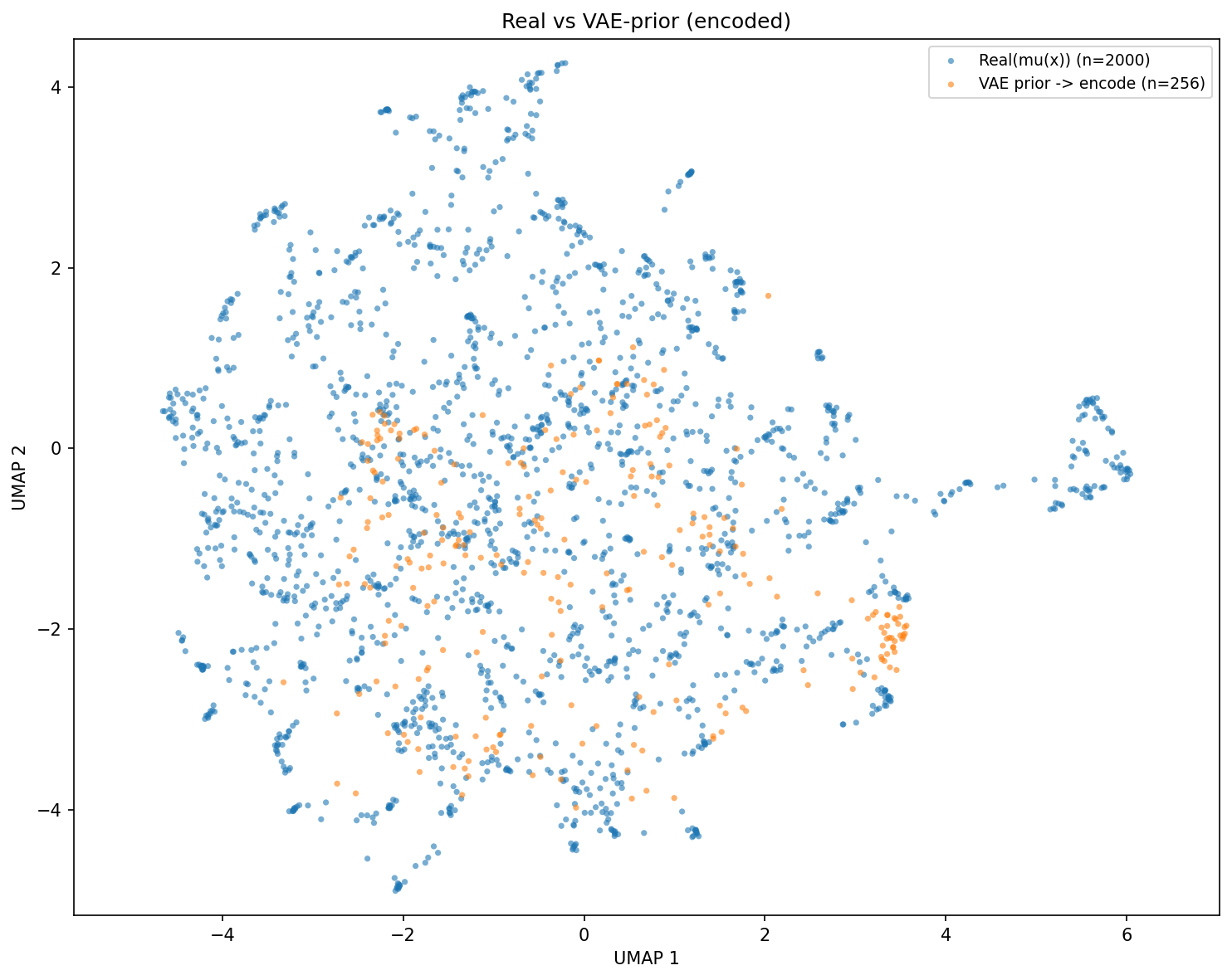}
  \caption{Real vs VAE prior.}
\end{subfigure}\hfill
\begin{subfigure}[t]{0.32\linewidth}
  \centering
  \includegraphics[width=\linewidth]{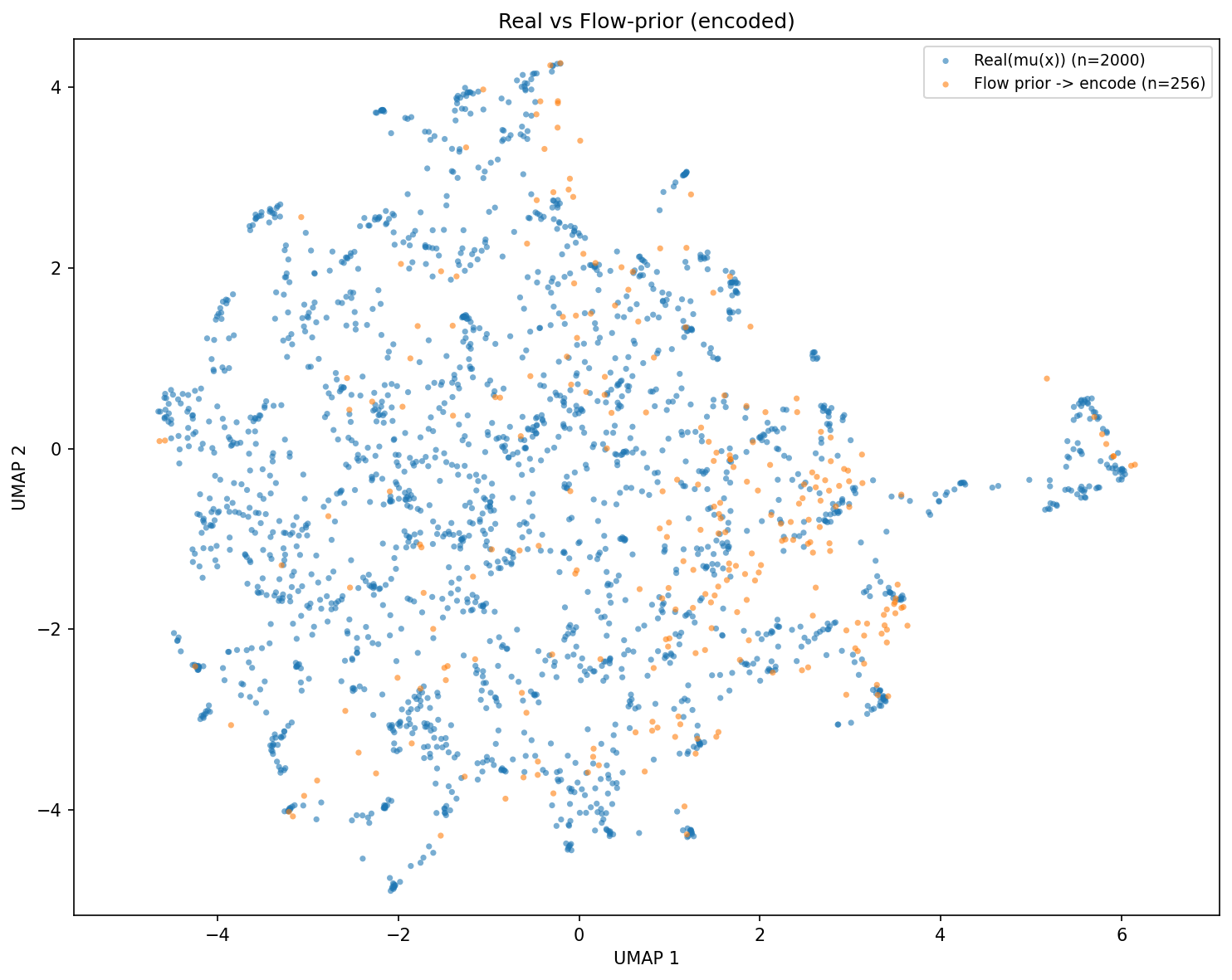}
  \caption{Real vs flow prior.}
\end{subfigure}
\caption{\textbf{UMAP of posterior-mean latent embeddings} $\mu_\phi(x)$ for real molecules (fit set) and for generated molecules after re-encoding.
Flow-prior samples align more tightly with the support of real embeddings compared to VAE-prior samples, consistent with improved validity and reduced FID-FP in Table~\ref{tab:ablation_generation_combined}.}
\label{fig:umap_smiles_a1}
\end{figure*}

\subsubsection{Flow matching ablation}
\label{sec:ablation_a1_smiles}

We begin by comparing unconditional sampling from a standard VAE Gaussian prior against sampling from a learned flow-matching prior, holding the decoder and training data fixed.
In both cases, latent samples are decoded without surrogate guidance ($\gamma=0$), allowing us to attribute differences in generation quality directly to the choice of latent prior.
Table~\ref{tab:ablation_generation_combined} summarizes the resulting generation statistics.
Replacing the Gaussian prior with a flow-matching prior improves validity and substantially reduces distributional discrepancy, as measured by fingerprint Fr\'echet distance and descriptor-level KL divergence.
Uniqueness and novelty among valid samples remain high in both settings, indicating that gains are not achieved by mode collapse.
For reference, we also report reconstruction from posterior means, which serves as an approximate upper bound on validity but does not constitute an unconditional generator.

The structure of the generated samples in latent space further clarifies these trends.
Figure~\ref{fig:umap_smiles_a1} visualizes posterior-mean latent embeddings of real molecules alongside embeddings of generated samples after re-encoding.
Samples drawn from the flow prior concentrate more tightly within the support of the real-data manifold, whereas samples from the VAE prior spread into low-density regions.
This qualitative alignment is consistent with the improvements in validity and distributional metrics observed in Table~\ref{tab:ablation_generation_combined}, and supports the role of the flow prior as a manifold-aware regularizer for latent sampling.

\subsubsection{Molecular representation ablation}
\label{sec:ablation_a2_smiles_vs_selfies}

We then investigate how the choice of molecular representation interacts with the learned flow prior.
Using the same VAE+flow architecture, we compare models trained on SMILES strings against those trained on SELFIES, again sampling unconditionally from the flow prior.
As shown in Table~\ref{tab:ablation_generation_combined}, SELFIES yields perfect syntactic validity while preserving high uniqueness and novelty among generated molecules.
Scaffold diversity remains comparable across representations, while distributional metrics reveal a modest trade-off: SELFIES improves descriptor-level KL divergence but exhibits slightly higher fingerprint Fr\'echet distance.
This behavior reflects the well-known tension between enforcing syntactic constraints and matching fine-grained distributional statistics.

Figure~\ref{fig:umap_a2_smiles_vs_selfies} provides a latent-space perspective on this comparison.
For both representations, flow-prior samples re-encode to regions overlapping with the real-data manifold, but SELFIES exhibits tighter coverage and fewer outliers.
Importantly, these differences arise despite identical training procedures and latent dynamics, underscoring the impact of representation choice on the effective support of the generative model.

While unconditional generation is not the primary focus of MoltenFlow, these results provide essential context for the optimizer-mode experiments, where the same flow prior regularizes guided trajectories and helps prevent drift away from the data manifold.

\begin{table*}[t]
\centering
\small
\setlength{\tabcolsep}{4.1pt}
\renewcommand{\arraystretch}{1.05}
\caption{\textbf{Unconditional generation ablations (no surrogate guidance).}
\textbf{A1} compares the VAE prior vs.\ flow-matching (FM) prior under SMILES.
\textbf{A2} compares SMILES vs.\ SELFIES under the FM prior.
FD-FP is Fr\'echet distance in fingerprint embedding space; Avg KL is the mean KL divergence over RDKit descriptors.}
\label{tab:ablation_generation_combined}
\begin{tabular}{@{}llcccccc@{}}
\toprule
Rep. & Prior & Valid(\%)$\uparrow$ & Unique(\%)$\uparrow$ & Novel(\%)$\uparrow$ & Scaf.\ div.$\uparrow$ & Avg KL$\downarrow$ & FD-FP$\downarrow$ \\
\midrule
SMILES   & VAE prior              & 5.76   & 98.44 & 99.47 & 0.944 & 0.299 & 13.23 \\
SMILES   & Flow prior (FM)        & 9.56   & 97.91 & 99.04 & 0.700 & 0.191 & 8.15  \\
SELFIES  & Flow prior (FM)        & 100.00 & 99.07 & 99.66 & 0.694 & 0.161 & 10.72 \\
\midrule
SMILES   & VAE post.\ ($\mu_\phi(x)$; ref.) & 26.07 & 78.02 & 71.58 & 0.438 & 0.207 & 4.67 \\
\bottomrule
\end{tabular}
\end{table*}

\section{Conclusion}
\label{sec:conclusion}

We introduced \textbf{MoltenFlow}, a latent-space framework for molecular generation and optimization that combines property-oriented latent space organization, flow-based generative modeling, and surrogate-based guidance.
By unifying generation and optimization within a single latent-space formulation, MoltenFlow enables stable and controllable multi-objective molecular design.

Empirically, we evaluated MoltenFlow on multi-objective optimization tasks derived from ZINC250K, focusing on maximizing drug-likeness (QED) while minimizing synthetic accessibility (SA).
Across both budgeted optimization and Pareto-front advancement experiments, MoltenFlow achieved faster and larger Pareto improvements than latent-space Bayesian optimization and unregularized gradient ascent under identical oracle budgets.
The guidance strength parameter $\gamma$ provided a clear and interpretable control over the trade-off between property improvement and structural fidelity, with an intermediate regime yielding substantial gains before over-optimization.
In unguided generation, learning a latent flow prior improved validity and distributional alignment relative to the VAE prior, with SELFIES further ensuring syntactic robustness.

Across these studies, MoltenFlow demonstrates that guided latent flows can reliably improve multi-objective performance while providing explicit control over optimization aggressiveness.
Rather than enforcing validity or diversity implicitly, the framework exposes these trade-offs through interpretable parameters, enabling principled navigation of the latent design space.
These results indicate that combining property-organized latent representations with learned generative priors can substantially improve the reliability and transparency of latent-space molecular optimization.

\paragraph{Limitations.}
MoltenFlow relies on surrogate property models and is therefore sensitive to surrogate accuracy and calibration, particularly under strong guidance. The current formulation operates on 2D molecular representations and does not capture three-dimensional structure or conformational effects. Performance is also sensitive to inference-time hyperparameters such as integration start time and noise scale. Additionally, our evaluation focuses on two proxy objectives (QED and SA), which do not capture the full complexity of molecular design tasks. Finally, while MoltenFlow provides controlled local optimization dynamics, it does not offer formal guarantees of global optimality.

\paragraph{Future Work.}
Incorporating uncertainty-aware or ensemble-based surrogate models could improve robustness and enable risk-sensitive optimization.
Adaptive schemes for selecting integration start time and noise scale—potentially informed by local latent density, surrogate uncertainty, or trajectory diagnostics—could further reduce sensitivity to inference-time hyperparameters.
Extending the framework to incorporate additional molecular properties beyond QED and SA, three-dimensional representations, and physics-informed objectives would broaden applicability to structure-sensitive design tasks. Evaluating on materials science benchmarks and integrating with experimental feedback loops are also important directions that could enable closed-loop discovery pipelines that combine latent optimization with real-world validation.


\bibliography{iclr2026_conference}

@article{gomez2018automatic,
  title={Automatic Chemical Design Using a Data-Driven Continuous Representation of Molecules},
  author={G{\'o}mez-Bombarelli, Rafael and Wei, Jennifer N. and Duvenaud, David and others},
  journal={ACS Central Science},
  volume={4},
  number={2},
  pages={268--276},
  year={2018}
}

@misc{kusner2017grammar,
      title={Grammar Variational Autoencoder}, 
      author={Matt J. Kusner and Brooks Paige and José Miguel Hern{\'a}ndez-Lobato},
      year={2017},
      eprint={1703.01925},
      archivePrefix={arXiv},
      primaryClass={stat.ML},
      url={https://arxiv.org/abs/1703.01925}, 
}

@misc{jin2018junction,
      title={Junction Tree Variational Autoencoder for Molecular Graph Generation}, 
      author={Wengong Jin and Regina Barzilay and Tommi Jaakkola},
      year={2019},
      eprint={1802.04364},
      archivePrefix={arXiv},
      primaryClass={cs.LG},
      url={https://arxiv.org/abs/1802.04364}, 
}

@article{simonovsky2018graphvae,
  title={GraphVAE: Towards Generation of Small Graphs Using Variational Autoencoders},
  author={Simonovsky, Martin and Komodakis, Nikos},
  journal={ICANN},
  year={2018}
}

@article{olivecrona2017molecular,
  title={Molecular De Novo Design through Deep Reinforcement Learning},
  author={Olivecrona, Marcus and Blaschke, Thomas and Engkvist, Ola and Chen, Hongming},
  journal={Journal of Cheminformatics},
  year={2017}
}

@article{zhou2019optimization,
  title   = {Optimization of Molecules via Deep Reinforcement Learning},
  author  = {Zhou, Zhenpeng and Kearnes, Steven and Li, Li and Zare, Richard N. and Riley, Patrick},
  journal = {Scientific Reports},
  volume  = {9},
  number  = {1},
  pages   = {10752},
  year    = {2019},
  doi     = {10.1038/s41598-019-47148-x},
  url     = {https://doi.org/10.1038/s41598-019-47148-x}
}

@article{griffiths2020constrained,
  title={Constrained Bayesian Optimization for Automatic Chemical Design},
  author={Griffiths, Ryan-Rhys and others},
  journal={Chemical Science},
  year={2020}
}

@article{hoogeboom2022equivariant,
  title={Equivariant Diffusion for Molecule Generation in 3D},
  author={Hoogeboom, Emiel and others},
  journal={ICML},
  year={2022}
}

@article{xu2022geodiff,
  title={GeoDiff: A Geometric Diffusion Model for Molecular Conformation Generation},
  author={Xu, Minkai and others},
  journal={ICLR},
  year={2022}
}

@article{dhariwal2021diffusion,
  title={Diffusion Models Beat GANs on Image Synthesis},
  author={Dhariwal, Prafulla and Nichol, Alexander},
  journal={NeurIPS},
  year={2021}
}

@misc{rezende2015variational,
      title={Variational Inference with Normalizing Flows}, 
      author={Danilo Jimenez Rezende and Shakir Mohamed},
      year={2016},
      eprint={1505.05770},
      archivePrefix={arXiv},
      primaryClass={stat.ML},
      url={https://arxiv.org/abs/1505.05770}, 
}

@misc{papamakarios2019normalizing,
      title={Normalizing Flows for Probabilistic Modeling and Inference}, 
      author={George Papamakarios and Eric Nalisnick and Danilo Jimenez Rezende and Shakir Mohamed and Balaji Lakshminarayanan},
      year={2021},
      eprint={1912.02762},
      archivePrefix={arXiv},
      primaryClass={stat.ML},
      url={https://arxiv.org/abs/1912.02762}, 
}

@misc{lipman2022flow,
      title={Flow Matching for Generative Modeling}, 
      author={Yaron Lipman and Ricky T. Q. Chen and Heli Ben-Hamu and Maximilian Nickel and Matt Le},
      year={2023},
      eprint={2210.02747},
      archivePrefix={arXiv},
      primaryClass={cs.LG},
      url={https://arxiv.org/abs/2210.02747}, 
}

@misc{albergo2023stochastic,
      title={Stochastic Interpolants: A Unifying Framework for Flows and Diffusions}, 
      author={Michael S. Albergo and Nicholas M. Boffi and Eric Vanden-Eijnden},
      year={2025},
      eprint={2303.08797},
      archivePrefix={arXiv},
      primaryClass={cs.LG},
      url={https://arxiv.org/abs/2303.08797}, 
}

@article{zang2020moflow,
  title={MoFlow: An Invertible Flow Model for Generating Molecular Graphs},
  author={Zang, Chengxi and Wang, Fei},
  journal={ICML},
  year={2020}
}

@misc{ho2022classifierfree,
  title         = {Classifier-Free Diffusion Guidance},
  author        = {Ho, Jonathan and Salimans, Tim},
  year          = {2022},
  eprint        = {2207.12598},
  archivePrefix = {arXiv},
  primaryClass  = {cs.LG},
  doi           = {10.48550/arXiv.2207.12598},
  url           = {https://arxiv.org/abs/2207.12598}
}

@inproceedings{chung2023dps,
  title     = {Diffusion Posterior Sampling for General Noisy Inverse Problems},
  author    = {Chung, Hyungjin and Kim, Jeongsol and McCann, Michael T. and Klasky, Marc L. and Ye, Jong Chul},
  booktitle = {International Conference on Learning Representations (ICLR)},
  year      = {2023},
  url       = {https://openreview.net/forum?id=OnD9zGAGT0k},
  eprint    = {2209.14687},
  archivePrefix = {arXiv},
  doi       = {10.48550/arXiv.2209.14687}
}

@misc{dao2023latentfm,
  title         = {Flow Matching in Latent Space},
  author        = {Dao, Quan and Phung, Hao and Nguyen, Binh and Tran, Anh},
  year          = {2023},
  eprint        = {2307.08698},
  archivePrefix = {arXiv},
  primaryClass  = {cs.CV},
  doi           = {10.48550/arXiv.2307.08698},
  url           = {https://arxiv.org/abs/2307.08698}
}

@article{tong2024otcfm,
  title   = {Improving and Generalizing Flow-Based Generative Models with Minibatch Optimal Transport},
  author  = {Tong, Alexander and Fatras, Kilian and Malkin, Nikolay and Huguet, Guillaume and Zhang, Yanlei and Rector-Brooks, Jarrid and Wolf, Guy and Bengio, Yoshua},
  journal = {Transactions on Machine Learning Research},
  year    = {2024},
  url     = {https://openreview.net/forum?id=CD9Snc73AW},
  eprint  = {2302.00482},
  archivePrefix = {arXiv},
  doi     = {10.48550/arXiv.2302.00482}
}

@inproceedings{liu2023rectifiedflow,
  title     = {Flow Straight and Fast: Learning to Generate and Transfer Data with Rectified Flow},
  author    = {Liu, Xingchao and Gong, Chengyue and Liu, Qiang},
  booktitle = {International Conference on Learning Representations (ICLR)},
  year      = {2023},
  url       = {https://openreview.net/forum?id=XVjTT1nw5z},
  eprint    = {2209.03003},
  archivePrefix = {arXiv},
  doi       = {10.48550/arXiv.2209.03003}
}

@inproceedings{chen2018neuralode,
  title     = {Neural Ordinary Differential Equations},
  author    = {Chen, Ricky T. Q. and Rubanova, Yulia and Bettencourt, Jesse and Duvenaud, David},
  booktitle = {Advances in Neural Information Processing Systems (NeurIPS)},
  year      = {2018},
  url       = {https://arxiv.org/abs/1806.07366},
  eprint    = {1806.07366},
  archivePrefix = {arXiv},
  doi       = {10.48550/arXiv.1806.07366}
}

@inproceedings{grathwohl2019ffjord,
  title     = {FFJORD: Free-Form Continuous Dynamics for Scalable Reversible Generative Models},
  author    = {Grathwohl, Will and Chen, Ricky T. Q. and Bettencourt, Jesse and Sutskever, Ilya and Duvenaud, David},
  booktitle = {International Conference on Learning Representations (ICLR)},
  year      = {2019},
  url       = {https://openreview.net/forum?id=rJxgknCcK7},
  eprint    = {1810.01367},
  archivePrefix = {arXiv},
  doi       = {10.48550/arXiv.1810.01367}
}

@inproceedings{ho2020ddpm,
  title     = {Denoising Diffusion Probabilistic Models},
  author    = {Ho, Jonathan and Jain, Ajay and Abbeel, Pieter},
  booktitle = {Advances in Neural Information Processing Systems (NeurIPS)},
  year      = {2020},
  url       = {https://proceedings.neurips.cc/paper/2020/hash/4c5bcfec8584af0d967f1ab10179ca4b-Abstract.html}
}

@inproceedings{song2021sde,
  title     = {Score-Based Generative Modeling through Stochastic Differential Equations},
  author    = {Song, Yang and Sohl-Dickstein, Jascha and Kingma, Diederik P. and Kumar, Abhishek and Ermon, Stefano and Poole, Ben},
  booktitle = {International Conference on Learning Representations (ICLR)},
  year      = {2021},
  url       = {https://arxiv.org/abs/2011.13456},
  eprint    = {2011.13456},
  archivePrefix = {arXiv},
  doi       = {10.48550/arXiv.2011.13456}
}

@misc{dunn2024flowmol,
  title         = {Mixed Continuous and Categorical Flow Matching for 3D De Novo Molecule Generation},
  author        = {Dunn, Ian and Koes, David Ryan},
  year          = {2024},
  eprint        = {2404.19739},
  archivePrefix = {arXiv},
  primaryClass  = {cs.LG},
  doi           = {10.48550/arXiv.2404.19739},
  url           = {https://arxiv.org/abs/2404.19739}
}

@misc{zeng2025propmolflow,
  title         = {PropMolFlow: Property-guided Molecule Generation with Geometry-Complete Flow Matching},
  author        = {Zeng, Cheng and Jin, Jirui and Karypis, George and Transtrum, Mark and Tadmor, Ellad B. and Hennig, Richard G. and Roitberg, Adrian and Martiniani, Stefano and Liu, Mingjie},
  year          = {2025},
  eprint        = {2505.21469},
  archivePrefix = {arXiv},
  primaryClass  = {cs.LG},
  doi           = {10.48550/arXiv.2505.21469},
  url           = {https://arxiv.org/abs/2505.21469}
}

@misc{irwin2024semlaflow,
  title         = {Efficient 3D Molecular Generation with Flow Matching and Scale Optimal Transport},
  author        = {Irwin, Ross and Tibo, Alessandro and Janet, Jon Paul and Olsson, Simon},
  year          = {2024},
  eprint        = {2406.07266},
  archivePrefix = {arXiv},
  primaryClass  = {cs.LG},
  doi           = {10.48550/arXiv.2406.07266},
  url           = {https://arxiv.org/abs/2406.07266}
}

@article{weininger1988smiles,
  title={SMILES, a chemical language and information system. 1. Introduction to methodology and encoding rules},
  author={Weininger, David},
  journal={Journal of Chemical Information and Computer Sciences},
  volume={28},
  number={1},
  pages={31--36},
  year={1988},
  publisher={ACS}
}

@article{mcinnes2018umap,
  title={UMAP: Uniform Manifold Approximation and Projection for Dimension Reduction},
  author={McInnes, Leland and Healy, John and Melville, James},
  journal={arXiv preprint arXiv:1802.03426},
  year={2018}
}

@misc{eckmann2022limolatentinceptionismtargeted,
      title={LIMO: Latent Inceptionism for Targeted Molecule Generation}, 
      author={Peter Eckmann and Kunyang Sun and Bo Zhao and Mudong Feng and Michael K. Gilson and Rose Yu},
      year={2022},
      eprint={2206.09010},
      archivePrefix={arXiv},
      primaryClass={cs.LG},
      url={https://arxiv.org/abs/2206.09010}, 
}

@misc{lin2025tfgflowtrainingfreeguidancemultimodal,
      title={TFG-Flow: Training-free Guidance in Multimodal Generative Flow}, 
      author={Haowei Lin and Shanda Li and Haotian Ye and Yiming Yang and Stefano Ermon and Yitao Liang and Jianzhu Ma},
      year={2025},
      eprint={2501.14216},
      archivePrefix={arXiv},
      primaryClass={cs.LG},
      url={https://arxiv.org/abs/2501.14216}, 
}

@article{sindt2026screening,
title = {Structure-based virtual screening of ultra-large chemical spaces: Advances and pitfalls},
journal = {European Journal of Medicinal Chemistry},
volume = {305},
pages = {118576},
year = {2026},
issn = {0223-5234},
doi = {https://doi.org/10.1016/j.ejmech.2026.118576},
url = {https://www.sciencedirect.com/science/article/pii/S0223523426000218},
author = {François Sindt and Didier Rognan}
}

@misc{tripp2020bayesian,
      title={Sample-Efficient Optimization in the Latent Space of Deep Generative Models via Weighted Retraining}, 
      author={Austin Tripp and Erik Daxberger and José Miguel Hernández-Lobato},
      year={2020},
      eprint={2006.09191},
      archivePrefix={arXiv},
      primaryClass={cs.LG},
      url={https://arxiv.org/abs/2006.09191}, 
}

@misc{wei2024chemflow,
      title={Navigating Chemical Space with Latent Flows}, 
      author={Guanghao Wei and Yining Huang and Chenru Duan and Yue Song and Yuanqi Du},
      year={2024},
      eprint={2405.03987},
      archivePrefix={arXiv},
      primaryClass={cs.LG},
      url={https://arxiv.org/abs/2405.03987}, 
}

@misc{jin2025molguidance,
      title={MolGuidance: Advanced Guidance Strategies for Conditional Molecular Generation with Flow Matching}, 
      author={Jirui Jin and Cheng Zeng and Pawan Prakash and Ellad B. Tadmor and Adrian Roitberg and Richard G. Hennig and Stefano Martiniani and Mingjie Liu},
      year={2025},
      eprint={2512.12198},
      archivePrefix={arXiv},
      primaryClass={cs.LG},
      url={https://arxiv.org/abs/2512.12198}, 
}

@misc{maus2023lolbo,
      title={Local Latent Space Bayesian Optimization over Structured Inputs}, 
      author={Natalie Maus and Haydn T. Jones and Juston S. Moore and Matt J. Kusner and John Bradshaw and Jacob R. Gardner},
      year={2023},
      eprint={2201.11872},
      archivePrefix={arXiv},
      primaryClass={cs.LG},
      url={https://arxiv.org/abs/2201.11872}, 
}

@misc{lee2025nfbo,
      title={Latent Bayesian Optimization via Autoregressive Normalizing Flows}, 
      author={Seunghun Lee and Jinyoung Park and Jaewon Chu and Minseo Yoon and Hyunwoo J. Kim},
      year={2025},
      eprint={2504.14889},
      archivePrefix={arXiv},
      primaryClass={cs.LG},
      url={https://arxiv.org/abs/2504.14889}, 
}

@misc{sabour2025alignflowscalingcontinuoustime,
      title={Align Your Flow: Scaling Continuous-Time Flow Map Distillation}, 
      author={Amirmojtaba Sabour and Sanja Fidler and Karsten Kreis},
      year={2025},
      eprint={2506.14603},
      archivePrefix={arXiv},
      primaryClass={cs.CV},
      url={https://arxiv.org/abs/2506.14603}, 
}

@misc{akiba2019optuna,
      title={Optuna: A Next-generation Hyperparameter Optimization Framework}, 
      author={Takuya Akiba and Shotaro Sano and Toshihiko Yanase and Takeru Ohta and Masanori Koyama},
      year={2019},
      eprint={1907.10902},
      archivePrefix={arXiv},
      primaryClass={cs.LG},
      url={https://arxiv.org/abs/1907.10902}, 
}

@InProceedings{hutter2014efficient,
  title = 	 {An Efficient Approach for Assessing Hyperparameter Importance},
  author = 	 {Hutter, Frank and Hoos, Holger and Leyton-Brown, Kevin},
  booktitle = 	 {Proceedings of the 31st International Conference on Machine Learning},
  pages = 	 {754--762},
  year = 	 {2014},
  editor = 	 {Xing, Eric P. and Jebara, Tony},
  volume = 	 {32},
  series = 	 {Proceedings of Machine Learning Research},
  address = 	 {Bejing, China},
  month = 	 {22--24 Jun},
  publisher =    {PMLR},
  url = 	 {https://proceedings.mlr.press/v32/hutter14.html},
}

@inproceedings{bonilla2007multi,
 author = {Bonilla, Edwin V and Chai, Kian and Williams, Christopher},
 booktitle = {Advances in Neural Information Processing Systems},
 editor = {J. Platt and D. Koller and Y. Singer and S. Roweis},
 pages = {},
 publisher = {Curran Associates, Inc.},
 title = {Multi-task Gaussian Process Prediction},
 url = {https://proceedings.neurips.cc/paper_files/paper/2007/file/66368270ffd51418ec58bd793f2d9b1b-Paper.pdf},
 volume = {20},
 year = {2007}
}

@misc{daulton2021parallel,
      title={Parallel Bayesian Optimization of Multiple Noisy Objectives with Expected Hypervolume Improvement}, 
      author={Samuel Daulton and Maximilian Balandat and Eytan Bakshy},
      year={2021},
      eprint={2105.08195},
      archivePrefix={arXiv},
      primaryClass={cs.LG},
      url={https://arxiv.org/abs/2105.08195}, 
}

@misc{balandat2020botorch,
      title={BoTorch: A Framework for Efficient Monte-Carlo Bayesian Optimization}, 
      author={Maximilian Balandat and Brian Karrer and Daniel R. Jiang and Samuel Daulton and Benjamin Letham and Andrew Gordon Wilson and Eytan Bakshy},
      year={2020},
      eprint={1910.06403},
      archivePrefix={arXiv},
      primaryClass={cs.LG},
      url={https://arxiv.org/abs/1910.06403}, 
}

@article{mann1947test,
  title={On a test of whether one of two random variables is stochastically larger than the other},
  author={Mann, Henry B and Whitney, Donald R},
  journal={The Annals of Mathematical Statistics},
  volume={18},
  number={1},
  pages={50--60},
  year={1947},
  publisher={Institute of Mathematical Statistics}
}

@book{efron1993introduction,
  author    = {Efron, Bradley and Tibshirani, Robert J.},
  title     = {An Introduction to the Bootstrap},
  publisher = {Chapman and Hall},
  year      = {1993},
  address   = {New York},
  series    = {Monographs on Statistics and Applied Probability},
  volume    = {57},
  isbn      = {0412042312}
}

@article{irwin2012zinc,
  title   = {ZINC: A Free Tool to Discover Chemistry for Biology},
  author  = {Irwin, John J. and Sterling, Teague and Mysinger, Michael M. and Bolstad, Erin S. and Coleman, Ryan G.},
  journal = {Journal of Chemical Information and Modeling},
  volume  = {52},
  number  = {7},
  pages   = {1757--1768},
  year    = {2012},
  publisher = {American Chemical Society},
  doi     = {10.1021/ci3001277},
  url     = {https://doi.org/10.1021/ci3001277}
}

@article{krenn2020selfies,
   title={Self-referencing embedded strings (SELFIES): A 100% robust molecular string representation},
   volume={1},
   ISSN={2632-2153},
   url={http://dx.doi.org/10.1088/2632-2153/aba947},
   DOI={10.1088/2632-2153/aba947},
   number={4},
   journal={Machine Learning: Science and Technology},
   publisher={IOP Publishing},
   author={Krenn, Mario and Häse, Florian and Nigam, AkshatKumar and Friederich, Pascal and Aspuru-Guzik, Alan},
   year={2020},
   month=oct, pages={045024}
}

@manual{rdkit,
  author = {Greg Landrum},
  title = {{RDKit}: Open-source cheminformatics},
  url = {https://www.rdkit.org/},
  year = {various}
}

@article{bickerton2012quantifying,
  title     = {Quantifying the Chemical Beauty of Drugs},
  author    = {Bickerton, George R. and Paolini, Giulio V. and Besnard, J{\'e}r{\^o}me and Muresan, Sorel and Hopkins, Andrew L.},
  journal   = {Nature Chemistry},
  volume    = {4},
  number    = {2},
  pages     = {90--98},
  year      = {2012},
  month     = jan,
  doi       = {10.1038/nchem.1243},
  pmid      = {22270643},
  pmcid     = {PMC3524573},
  publisher = {Nature Publishing Group}
}

@article{ertl2009estimation,
  title     = {Estimation of Synthetic Accessibility Score of Drug-like Molecules Based on Molecular Complexity and Fragment Contributions},
  author    = {Ertl, Peter and Schuffenhauer, Ansgar},
  journal   = {Journal of Cheminformatics},
  volume    = {1},
  number    = {1},
  pages     = {8},
  year      = {2009},
  month     = jun,
  doi       = {10.1186/1758-2946-1-8},
  pmid      = {20298526},
  pmcid     = {PMC3225829},
  publisher = {BioMed Central}
}

@misc{you2019graph,
      title={Graph Convolutional Policy Network for Goal-Directed Molecular Graph Generation}, 
      author={Jiaxuan You and Bowen Liu and Rex Ying and Vijay Pande and Jure Leskovec},
      year={2019},
      eprint={1806.02473},
      archivePrefix={arXiv},
      primaryClass={cs.LG},
      url={https://arxiv.org/abs/1806.02473}, 
}

@article{blank2020pymoo,
   title={Pymoo: Multi-Objective Optimization in Python},
   volume={8},
   ISSN={2169-3536},
   url={http://dx.doi.org/10.1109/ACCESS.2020.2990567},
   DOI={10.1109/access.2020.2990567},
   journal={IEEE Access},
   publisher={Institute of Electrical and Electronics Engineers (IEEE)},
   author={Blank, Julian and Deb, Kalyanmoy},
   year={2020},
   pages={89497–89509}
}

@article{bemis1996,
  author = {Bemis, Guy W. and Murcko, Mark A.},
  title = {The properties of known drugs. 1. Molecular frameworks},
  journal = {Journal of Medicinal Chemistry},
  volume = {39},
  number = {15},
  pages = {2887--2893},
  year = {1996},
  doi = {10.1021/jm9602928},
  URL = {https://doi.org},
  eprint = {https://doi.org}
}
\bibliographystyle{iclr2026_conference}

\clearpage
\appendix
\section{Extended Related Work}
\label{sec:extended-related-work}

\subsection{Molecular representations and validity.}
String-based molecular representations such as SMILES \citep{weininger1988smiles} are compact but prone to syntactic invalidity.
SELFIES provides a robust alternative that guarantees syntactic validity by construction \citep{krenn2020selfies}.
Dimensionality reduction techniques such as UMAP are commonly used to visualize latent-space structure and assess coverage \citep{mcinnes2018umap}.

\subsection{Molecular latent representation learning.}
A central theme in molecular generative modeling is the use of continuous latent representations to capture chemical variation in a form amenable to interpolation and optimization.
Early work with SMILES-based variational autoencoders demonstrated that latent spaces learned from molecular strings encode meaningful structure and permit gradient-based exploration \cite{gomez2018automatic,kusner2017grammar}.
Subsequent extensions incorporated graph-based representations and structured decoders to improve validity and expressiveness \cite{jin2018junction,simonovsky2018graphvae}.
Despite these advances, latent representations learned purely for reconstruction are not inherently organized for inverse design, and unconstrained traversal in such spaces often leads to invalid molecules or degraded diversity when aggressively optimizing properties.

\subsection{Goal-directed molecular optimization.} 
A large body of work addresses the problem of generating molecules that optimize one or more target properties.
Reinforcement learning approaches formulate molecule construction as a sequential decision process, allowing direct optimization of reward functions \cite{olivecrona2017molecular,zhou2019optimization}.
While flexible, these methods are sensitive to reward shaping and frequently require careful tuning to avoid mode collapse.
Bayesian optimization has also been applied in learned latent spaces as a sample-efficient alternative \cite{griffiths2020constrained,tripp2020bayesian}, with recent methods such as LOL-BO introducing trust-region strategies to stabilize search \cite{maus2023lolbo}.
However, latent-space BO methods typically rely on fitting an additional surrogate over the latent space and can struggle with sample efficiency as latent dimensionality increases.

\subsection{Diffusion and flow-based generative models.}
Diffusion-based models have achieved state-of-the-art performance in molecular generation, particularly when incorporating geometric inductive biases \citep{hoogeboom2022equivariant,xu2022geodiff}.
From a broader perspective, diffusion can be understood through the score-based SDE framework, which unifies stochastic and deterministic sampling views and clarifies how guidance terms influence reverse-time dynamics \citep{ho2020ddpm,song2021sde}.
Classifier-guided and conditional diffusion techniques provide principled mechanisms for steering generation toward desired properties \citep{dhariwal2021diffusion}, but applying diffusion directly in molecular space is computationally expensive and controlling multi-objective optimization trajectories remains challenging.

\subsection{Flow matching as a generative prior.}
Flow-based generative models provide a complementary approach, learning continuous-time dynamics that transport samples from simple base distributions to complex target distributions \cite{rezende2015variational,papamakarios2019normalizing}.
Recent work on flow matching reframes this process as learning vector fields along predefined interpolation paths, avoiding explicit likelihood computation \cite{lipman2022flow,albergo2023stochastic}.
Applying flow matching in the latent space of a pretrained autoencoder yields a scalable analogue to latent diffusion \cite{dao2023latentfm}.
Several recent studies have applied flow matching to molecular generation, including MoFlow, FlowMol, SemlaFlow, and related frameworks \cite{zang2020moflow,dunn2024flowmol,irwin2024semlaflow}, with an emphasis on unconditional sampling or efficient generation rather than guided optimization.

\subsection{Normalizing flows and continuous-time models.}
Normalizing flows provide exact likelihood-based generative models through invertible transformations \citep{rezende2015variational,papamakarios2019normalizing}.
Continuous normalizing flows and neural ODEs extend this framework to continuous-time dynamics \citep{chen2018neuralode,grathwohl2019ffjord}.
Recent variants, including rectified flow and optimal-transport-based couplings, simplify training and reduce inference cost \citep{liu2023rectifiedflow,tong2024otcfm}.

\subsection{Guidance and conditional generation.}
Guidance mechanisms have been widely studied in diffusion and flow-based models, including classifier guidance, classifier-free guidance, and posterior sampling formulations \citep{dhariwal2021diffusion,chung2023dps,ho2022classifierfree}.
Recent work has explored training-free guidance in multimodal settings \citep{lin2025tfgflowtrainingfreeguidancemultimodal}.
These approaches typically modify the generative dynamics directly, whereas MoltenFlow uses surrogate-based gradients to guide a latent flow prior.

\subsection{Positioning this work.}
MoltenFlow builds on these three lines of work by combining property-organized latent representations with a learned latent flow prior and surrogate-based gradient guidance.
In contrast to latent-space Bayesian optimization, MoltenFlow avoids fitting a second surrogate over the latent space and instead exploits direct gradients from a differentiable property head.
Relative to recent flow-matching-based molecular generators such as ChemFlow, which learn energy or vector fields over a fixed latent space \cite{wei2024chemflow}, MoltenFlow explicitly reshapes the latent representation through auxiliary property supervision and uses the learned flow to regularize guided optimization trajectories.
This design enables controlled, sample-efficient multi-objective optimization while exposing an interpretable trade-off between improvement and feasibility.

\section{Molecular string representations (SMILES and SELFIES).}
\label{app:mol-representations}
MoltenFlow supports discrete molecular inputs $x$ provided as string representations.
By default we use SMILES \citep{weininger1988smiles}. For the representation ablation
(Section~\ref{sec:ablations}), we additionally train and sample models on SELFIES
\citep{krenn2020selfies}, a grammar-constrained string representation designed to
reduce syntactic invalidity under token-level perturbations.

\subsection{Tokenization.}
For SMILES, we tokenize at the character (or multi-character atom) level with special
tokens for padding and sequence boundaries (BOS/EOS), and truncate/pad sequences to
a fixed maximum length $L$.
For SELFIES, we tokenize by bracketed SELFIES symbols, yielding a sequence of discrete
tokens with the same BOS/EOS/padding convention and the same maximum length $L$.
All model architectures and hyperparameters are held fixed between SMILES and SELFIES
runs unless stated otherwise; only the tokenizer/vocabulary changes.

\subsection{Canonicalization and validity.}
For evaluation, generated strings are converted to canonical SMILES using RDKit
canonicalization. For SELFIES-trained models, generated SELFIES strings are first
decoded into SMILES (using the standard SELFIES decoder) and then canonicalized.
A sample is counted as \emph{valid} if it successfully parses into an RDKit molecule
after conversion/canonicalization; otherwise it is invalid.

\section{Objective Construction for Guided Latent Dynamics}
\label{app:objective_construction}

This section provides explicit examples of how the scalar objective $\mathcal{J}(z;c)$ used for guided latent dynamics is constructed from surrogate predictions.
As described in \S\ref{sec:guidance}, $\mathcal{J}$ is interpreted as a loss function to be minimized, and maximization objectives are handled by appropriate sign conventions.

\paragraph{Target-based objectives.}
For objectives that specify desired target values for molecular properties, we define $\mathcal{J}$ as a weighted squared error between surrogate predictions and target values:
\begin{equation}
\mathcal{J}_{\text{target}}(z;c)
=
\sum_{i=1}^{P} w_i \left(f_{\psi,i}(z) - c_i\right)^2,
\end{equation}
where $f_{\psi,i}(z)$ denotes the surrogate prediction for property $i$, $c_i$ is the desired target value, and $w_i \ge 0$ controls the relative importance of each property.
This form encourages latent updates that move predictions toward specified targets.

\paragraph{Directional (maximize/minimize) objectives.}
For directional optimization, where properties are to be maximized or minimized without explicit targets, we define
\begin{equation}
\mathcal{J}_{\text{dir}}(z;c)
=
-\sum_{i=1}^{P} s_i \, f_{\psi,i}(z),
\end{equation}
where $s_i \in \{+1,-1\}$ encodes the optimization direction for property $i$.
Specifically, $s_i = +1$ corresponds to maximizing property $i$, while $s_i = -1$ corresponds to minimizing it.
With this convention, gradient descent on $\mathcal{J}_{\text{dir}}$ implements the desired maximize/minimize behavior.

\paragraph{Encoding objectives via $c$.}
In practice, the conditioning vector $c$ specifies either target values (for $\mathcal{J}_{\text{target}}$) or directional indicators (for $\mathcal{J}_{\text{dir}}$), and may also include optional property weights.
Both formulations are compatible with the guided dynamics in Eq.~\ref{eq:flow_dynamics}.

\paragraph{Examples: QED and SA.}
In the experiments, we optimize for high drug-likeness (QED) and low synthetic accessibility (SA).
Using the directional formulation, this corresponds to setting
\[
s_{\text{QED}} = +1,
\qquad
s_{\text{SA}} = -1,
\]
so that gradient descent on $\mathcal{J}_{\text{dir}}$ increases QED while decreasing SA.
Alternatively, target-based objectives may be constructed by specifying desired target values $c_{\text{QED}}$ and $c_{\text{SA}}$ in $\mathcal{J}_{\text{target}}$.

\section{Model architecture and training details}
\label{app:arch}

\paragraph{VAE.}
We use a Transformer-based sequence VAE for string inputs (SMILES/SELFIES). The encoder is a Transformer encoder with $N_{\text{enc}}$ layers, model width $d_{\text{model}}$, $H$ attention heads, and feedforward width $d_{\text{ff}}$. The latent interface uses attention pooling to produce $K$ latent tokens of dimension $d$, parameterizing a diagonal Gaussian posterior $q_\phi(z\mid x)$. The decoder is a Transformer decoder with $N_{\text{dec}}$ layers that conditions on $z$ via cross-attention and is trained with teacher forcing.

\paragraph{Property surrogate.}
The property predictor $f_\psi$ operates on the pooled latent $\bar z$ and is implemented as an MLP with \emph{3} layers and hidden size \emph{1024}, with output heads for QED and SA. We bound outputs using sigmoid activation for QED and scaled sigmoid for SA to enforce $\mathrm{QED}\in[0,1]$ and $\mathrm{SA}\in[1,10]$.

\paragraph{Flow model.}
A Transformer-based architecture with 10 layers, hidden dimension 256, 8 attention heads, feedforward dimension 512, and dropout rate 0.1 parameterizes the velocity field.
Time conditioning uses a 128-dimensional sinusoidal embedding.
The model is trained for 100 epochs with batch size 1024 and learning rate $2 \times 10^{-4}$ using the conditional flow matching objective with optimal transport coupling \citep{lipman2022flow,tong2024otcfm}.

\paragraph{Training protocol.}
Stage (i) VAE pretraining uses objective Eq.~\ref{eq:vae-objective} for 150 epochs with optimizer AdamW, learning rate $1e-4$, batch size $256$. Stage (ii) property fine-tuning minimizes Eq.~\ref{eq:finetune-loss} with $\lambda=1$ for 150 epochs. Stage (iii) flow matching minimizes Eq.~\ref{eq:flow-loss} using $z_0\sim\mathcal{N}(0,I)$ and $z_1\sim q_{\phi'}(z\mid x)$ for 300 epochs, with learning rate $2e-4$ and batch size $1024$. For each stage, the epoch with the best validation loss is selected.

\section{Experiments Details}
\label{app:experiment-details}

\subsection{Hyperparameter Optimization for MoltenFlow}
\label{app:budgeted-hpo}

MoltenFlow hyperparameters were selected via Bayesian hyperparameter optimization using Optuna \citep{akiba2019optuna} with the TPE sampler.
We searched over 100 configurations, with each configuration evaluated over 3 random seeds at a reduced budget of $B = 50$ oracle calls.
The objective was mean HVI across seeds.

Table~\ref{tab:hpo_search_space} specifies the search space.
The guidance strength $\gamma$ and noise scale $\sigma$ were identified as the most important parameters via fANOVA importance analysis \citep{hutter2014efficient}.

\begin{table}[ht]
\centering
\small
\caption{Hyperparameter search space for MoltenFlow.}
\label{tab:hpo_search_space}
\begin{tabular}{llll}
\toprule
Parameter & Type & Range & Selected \\
\midrule
$\gamma$ (guidance strength) & Log-uniform & $[10, 50]$ & 36.07 \\
$\sigma$ (noise scale) & Uniform & $[0.4, 0.9]$ & 0.80 \\
steps (integration steps) & Integer & $[10, 50]$ & 12 \\
$t_{\text{start}}$ & Uniform & $[0.5, 0.9]$ & 0.89 \\
clip\_norm & Categorical & $\{\varnothing, 1, 5, 10\}$ & 5.0 \\
normalize\_gradient & Categorical & $\{\text{True}, \text{False}\}$ & True \\
\bottomrule
\end{tabular}
\end{table}

\begin{figure*}[ht]
\centering
\begin{subfigure}[t]{0.32\linewidth}
    \centering
    \includegraphics[width=\linewidth]{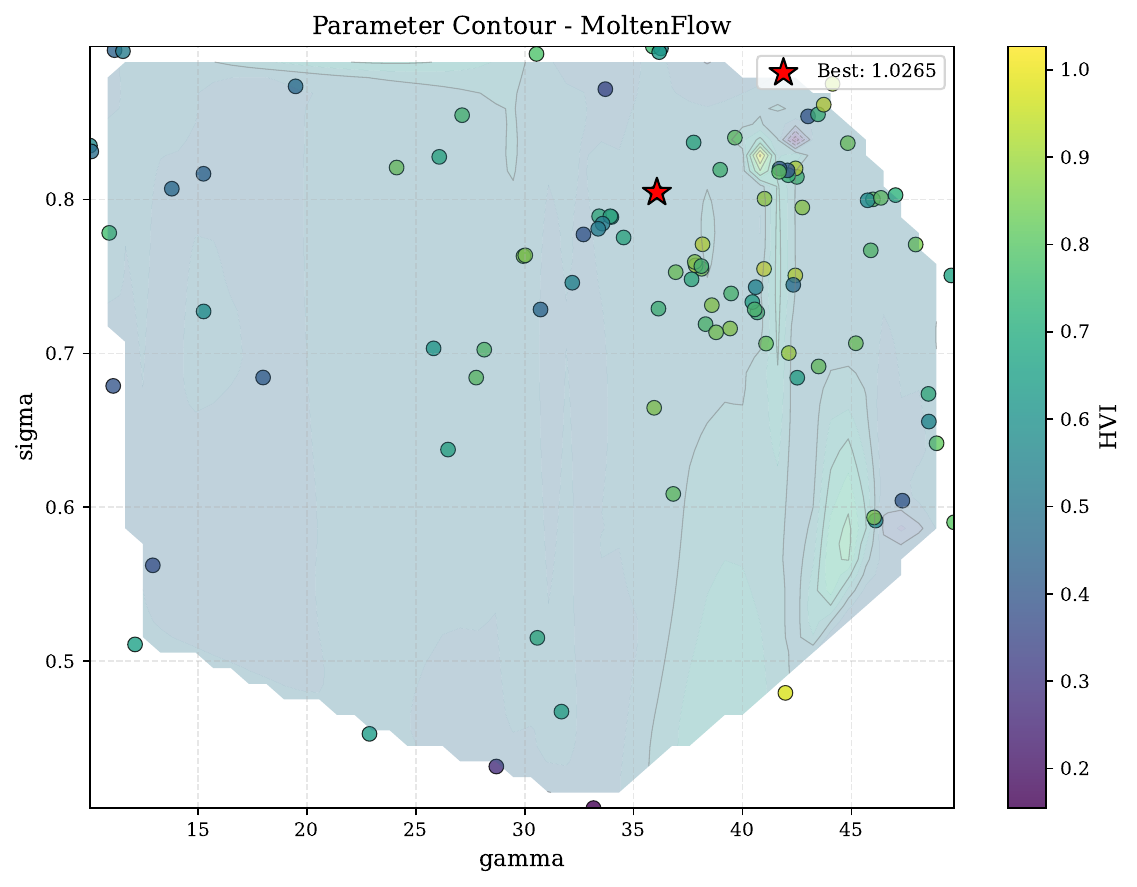}
    \caption{Objective contour}
    \label{fig:hpo-contour}
\end{subfigure}\hfill
\begin{subfigure}[t]{0.32\linewidth}
    \centering
    \includegraphics[width=\linewidth]{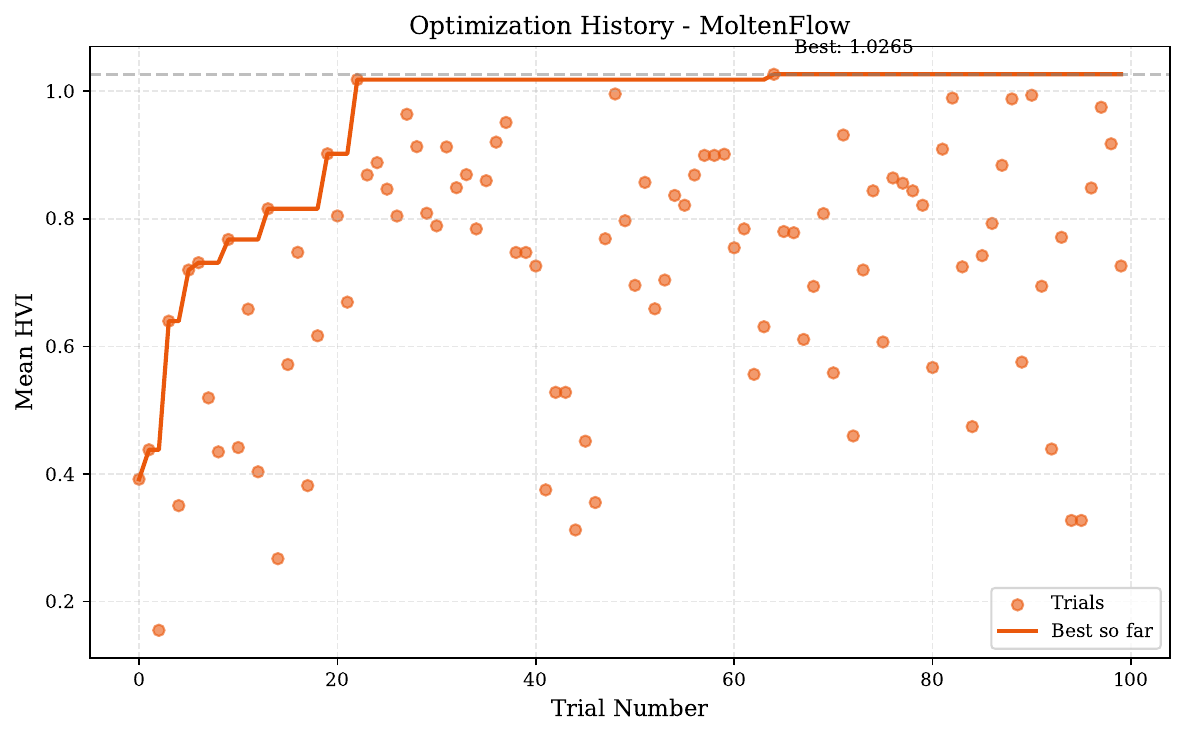}
    \caption{Optimization history.}
    \label{fig:hpo-history}
\end{subfigure}\hfill
\begin{subfigure}[t]{0.32\linewidth}
    \centering
    \includegraphics[width=\linewidth]{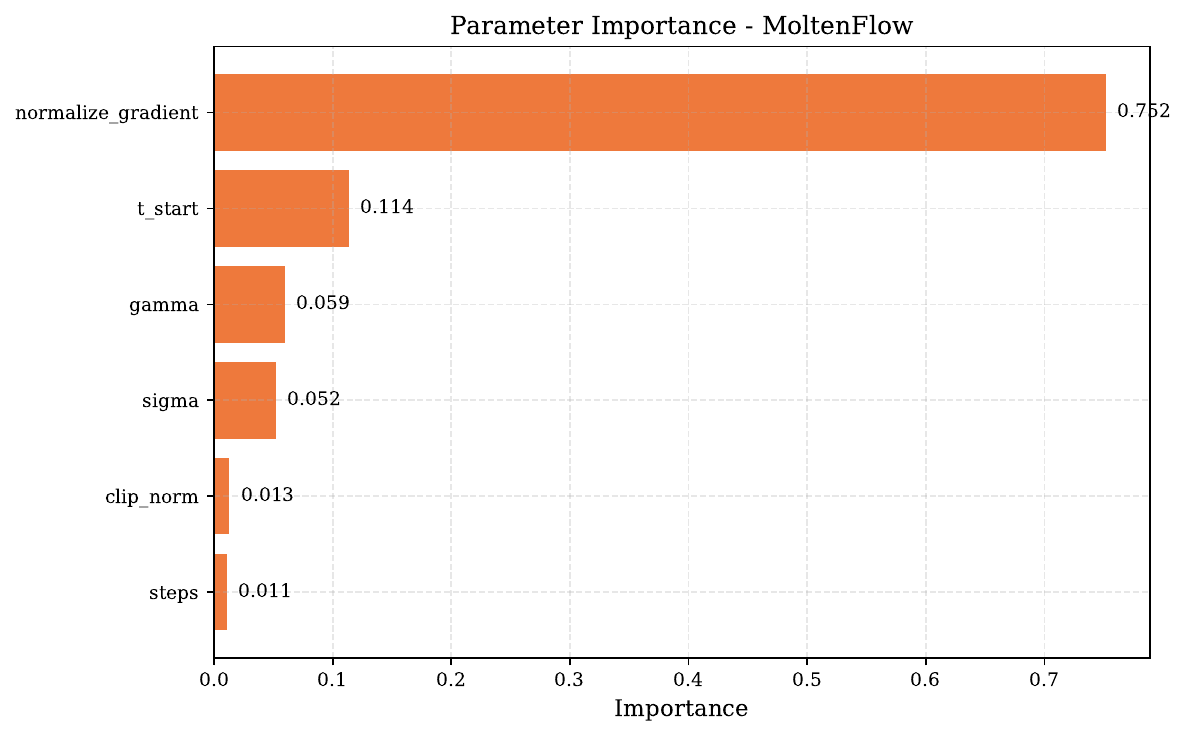}
    \caption{Hyperparameter importance}
    \label{fig:hpo-importance}
\end{subfigure}
\caption{\textbf{Hyperparameter optimization diagnostics.}
Left: contour of hypervolume improvement as a function of guidance strength $\gamma$ and noise scale $\sigma$.
Center: optimization history across trials.
Right: relative importance of hyperparameters for maximizing hypervolume improvement.
}
\label{fig:hpo-summary}
\end{figure*}

The best configuration achieved HVI $= 1.03$ (mean over 3 seeds) at trial 64, with $\gamma = 36.07$, $\sigma = 0.80$, steps $= 12$, $t_{\text{start}} = 0.89$, clip\_norm $= 5.0$, and normalize\_gradient $= \text{True}$.
Across all 100 trials, mean HVI was $0.72 \pm 0.20$, indicating substantial sensitivity to hyperparameter choice.

\subsection{Budgeted Multi-Objective Optimization}
\label{app:budgeted-details}

\subsubsection{Experimental Design}
The budgeted optimization experiment follows a similar experimental protocol for constrained oracle settings established in recent molecular optimization benchmarks \citep{maus2023lolbo,tripp2020bayesian}.
Each method operates under a fixed budget of $B = 100$ oracle evaluations, starting from an initial set of 10 randomly sampled molecules.
At each iteration, methods propose a single candidate molecule, which is decoded from the latent space and evaluated using RDKit oracles for QED and SA.
The evaluated molecule is added to the candidate pool, and the process repeats until the budget is exhausted.

\subsubsection{Baseline Implementations}

\paragraph{MoltenFlow.}
The full MoltenFlow pipeline uses the VAE, surrogate, and flow models described in Appendix~\ref{app:arch}.
At each iteration, a seed molecule is selected from the current Pareto front using diversity-weighted sampling, which balances exploitation of high-quality regions with exploration of undersampled areas.
Specifically, the selection probability for molecule $i$ is:
\begin{equation}
    p_i \propto w_{\text{pareto}}(i) \cdot \exp\left(-\lambda \cdot \text{sim}_{\max}(i)\right)
\end{equation}
where $w_{\text{pareto}}(i) = 2$ if $i$ is Pareto-optimal and 1 otherwise, $\text{sim}_{\max}(i)$ is the maximum Tanimoto similarity to recently optimized molecules, and $\lambda = 2$ is the diversity penalty.

Optimization proceeds by adding Gaussian noise ($\sigma = 0.80$) to the seed encoding and integrating the guided ODE for 12 steps starting from $t = 0.89$.
Gradients are clipped to norm 5.0 and normalized before scaling by $\gamma = 36.1$.

\paragraph{Gradient Ascent (No-Flow Ablation).}
The gradient ascent baseline removes the flow velocity term from the guided dynamics:
\begin{equation}
    \frac{dz}{dt} = \eta \cdot \nabla_z \mathcal{L}(z)
\end{equation}
where $\eta = 0.3$ is the step size and $\mathcal{L}(z)$ is the directional objective derived from surrogate predictions.
This ablation isolates the contribution of the flow prior to optimization performance.
The same diversity-weighted seed selection and noise injection ($\sigma = 0.2$) are applied. These parameters were not hyperparameter optimized.

\paragraph{Bayesian Optimization.}
We implement latent-space Bayesian optimization following \citet{gomez2018automatic} and \citet{tripp2020bayesian}.
Molecules are encoded using the VAE encoder, and latent representations are aggregated via mean pooling across tokens to produce 128-dimensional vectors.
For BO (MOGP), we fit a multi-output GP with a Mat\'ern-5/2 kernel and intrinsic coregionalization \citep{bonilla2007multi}.
For BO (2-GP), we fit independent GPs for QED and SA.
Acquisition is performed using qLogExpectedHypervolumeImprovement \citep{daulton2021parallel} with 10 restarts and 512 raw samples, optimized via L-BFGS-B.
We use BoTorch \citep{balandat2020botorch} for GP fitting and acquisition optimization.

\subsubsection{Statistical Analysis}

Performance differences were assessed using the Mann-Whitney U test \citep{mann1947test}, a non-parametric test appropriate for comparing distributions that may not be normally distributed.
All pairwise comparisons involving MoltenFlow yielded $p < 10^{-4}$, indicating statistically significant differences.
The comparison between the two BO variants was not significant ($p > 0.05$), suggesting that the multi-output GP formulation provides no advantage over independent GPs in this setting.

Confidence intervals were computed using the bootstrap percentile method with 1,000 resamples \citep{efron1993introduction}.
The 90\% confidence level was chosen to provide tighter visual bands while maintaining statistical rigor.

\subsubsection{Computational Resources}

All experiments were conducted on a Nvidia A100 GPU.
Per-run timings (Figure~\ref{fig:runtime_comparison}) reflect end-to-end optimization time including model inference, acquisition optimization (for BO), and oracle evaluation.
MoltenFlow's computational advantage stems from avoiding GP fitting ($\mathcal{O}(n^3)$ in the number of evaluated points) and acquisition optimization (requiring multiple gradient-based restarts).

\subsubsection{Extended Results}
\label{app:budget-extended-results}

Figure~\ref{fig:pareto_examples} shows representative Pareto fronts from the best and worst runs for each method.
MoltenFlow consistently produces dense fronts in the high-QED, low-SA region, while BO methods exhibit greater variability.

Figure~\ref{fig:runtime_comparison} compares runtime distributions across methods.
BO methods exhibit high variance in runtime due to the increasing cost of GP fitting as the evaluated set grows.

\begin{figure}[ht!]
\centering
\includegraphics[width=0.9\textwidth]{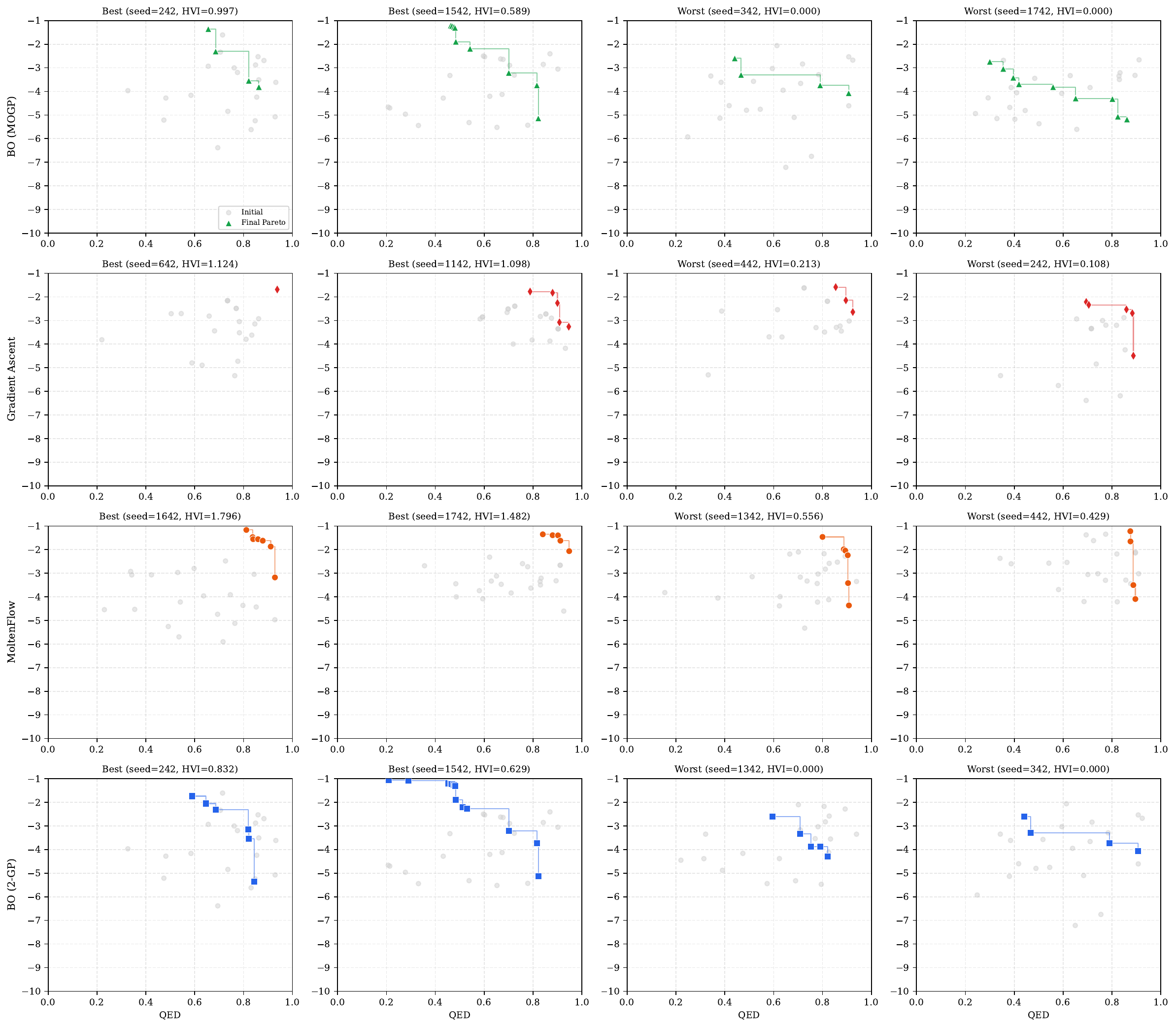}
\caption{Representative Pareto fronts from best (top) and worst (bottom) runs for each method. Gray points: initial molecules. Colored points: final Pareto front. MoltenFlow achieves consistent improvements; BO methods show higher variance.}
\label{fig:pareto_examples}
\end{figure}

\begin{figure}[ht!]
\centering
\includegraphics[width=0.6\linewidth]{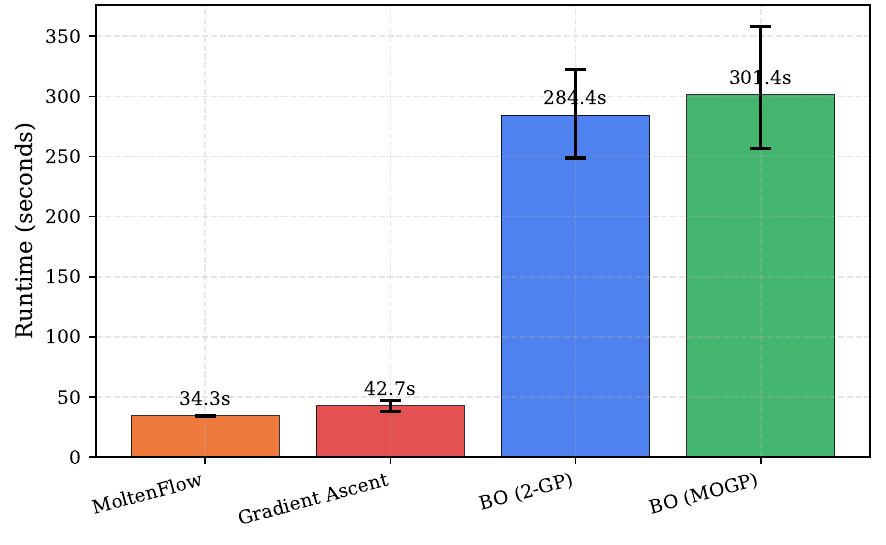}
\caption{Runtime comparison across methods ($n = 20$ seeds). Error bars: 90\% bootstrap CI. MoltenFlow is $\sim8\times$ faster than BO methods.}
\label{fig:runtime_comparison}
\end{figure}

\subsubsection{Hyperparameter Summary}

Table~\ref{tab:budgeted_hyperparams} summarizes all hyperparameters for the budgeted optimization experiment.

\begin{table}[ht]
\centering
\small
\caption{Hyperparameters for budgeted optimization experiments.}
\label{tab:budgeted_hyperparams}
\begin{tabular}{ll}
\toprule
\textbf{Parameter} & \textbf{Value} \\
\midrule
\multicolumn{2}{l}{\textit{Experimental Setup}} \\
Oracle budget ($B$) & 100 \\
Initial set size & 10 \\
Batch size & 1 \\
Seeds per method & 20 \\
Initialization & Random sampling \\
\midrule
\multicolumn{2}{l}{\textit{MoltenFlow}} \\
Guidance strength ($\gamma$) & 36.07 \\
Noise scale ($\sigma$) & 0.80 \\
Integration steps & 12 \\
Start time ($t_{\text{start}}$) & 0.89 \\
Gradient clipping & 5.0 \\
Gradient normalization & True \\
Seed selection & Diversity-weighted \\
Diversity penalty ($\lambda$) & 2.0 \\
Pareto weight & 2.0 \\
\midrule
\multicolumn{2}{l}{\textit{Gradient Ascent}} \\
Step size ($\eta$) & 0.3 \\
Noise scale ($\sigma$) & 0.2 \\
Steps & 10 \\
Seed selection & Diversity-weighted \\
\midrule
\multicolumn{2}{l}{\textit{Bayesian Optimization}} \\
GP kernel & Mat\'ern-5/2 \\
Latent aggregation & Mean pooling \\
Acquisition restarts & 10 \\
Raw samples & 512 \\
\midrule
\multicolumn{2}{l}{\textit{Hyperparameter Optimization}} \\
HPO method & TPE (Optuna) \\
Trials & 100 \\
Seeds per trial & 3 \\
Budget per trial & 50 \\
\midrule
\multicolumn{2}{l}{\textit{Evaluation}} \\
Confidence level & 0.90 \\
Bootstrap samples & 1,000 \\
Reference point & $(0.0, -10.0)$ \\
\bottomrule
\end{tabular}
\end{table}

\subsection{Multi-Objective Pareto Front Advancement}
\label{app:zinc-details}

\subsubsection{Dataset and Preprocessing}
We use the ZINC250K dataset \citep{irwin2012zinc}, containing approximately 250,000 drug-like molecules.
The dataset is split into training and test sets using scaffold-based splitting to ensure chemically diverse evaluation with an 81/9/10 scaffolding split.
Molecules are represented as SELFIES strings \citep{krenn2020selfies}, which guarantee syntactic validity upon decoding.
The vocabulary contains 111 tokens with a maximum sequence length of 128.

\subsubsection{Molecular Properties}
We compute three molecular properties using RDKit \citep{rdkit}:

\paragraph{Quantitative Estimate of Drug-likeness (QED).}
QED \citep{bickerton2012quantifying} is a composite score that quantifies drug-likeness based on eight molecular descriptors: molecular weight, octanol-water partition coefficient (ALogP), number of hydrogen bond donors, number of hydrogen bond acceptors, molecular polar surface area, number of rotatable bonds, number of aromatic rings, and number of structural alerts. Each descriptor is mapped to a desirability function based on the distributions observed in known drugs, and the final QED score is the geometric mean of all desirability values:
\begin{equation}
    \text{QED} = \exp\left(\frac{1}{n}\sum_{i=1}^{n} \ln d_i\right)
\end{equation}
where $d_i$ is the desirability function for descriptor $i$. QED $\in [0, 1]$, with higher values indicating more drug-like molecules.

\paragraph{Synthetic Accessibility Score (SA).}
SA \citep{ertl2009estimation} estimates the ease of chemical synthesis based on fragment contributions and molecular complexity:
\begin{equation}
    \text{SA} = \text{fragmentScore} - \text{complexityPenalty}
\end{equation}
where the fragment score is derived from the frequency of molecular fragments in the PubChem database (common fragments receive higher scores), and the complexity penalty accounts for features that complicate synthesis such as macrocycles, stereocenters, and spiro/bridged systems. SA $\in [1, 10]$, with lower values indicating easier synthesis.

\paragraph{Penalized LogP (pLogP).}
pLogP \citep{jin2018junction,you2019graph} combines lipophilicity with synthesizability and ring size constraints:
\begin{equation}
    \text{pLogP} = \log P - \text{SA} - \text{ringPenalty}
\end{equation}
where $\log P$ is the Crippen octanol-water partition coefficient, and $\text{ringPenalty} = \max(0, \text{largestRingSize} - 6)$ penalizes molecules with unusually large rings. This metric is commonly used in molecular optimization benchmarks as it captures a trade-off between desirable lipophilicity and practical constraints.

\subsubsection{Optimization Procedure}
Optimization initializes from latent encodings of test-set Pareto-optimal molecules (identified using surrogate predictions).
Gaussian noise ($\sigma = 0.8$) is added to each latent code.
Guided integration uses 12 Euler steps from $t_{\text{start}} = 0.89$ to $t = 1$ (per the parameters obtained in ~\ref{app:budgeted-hpo}) restricting the flow to the final 11\% of its trajectory where representations are near the data manifold.
The directional objective is derived from normalized gradients pointing toward improved Pareto regions.
Candidate molecules for optimization are selected from the test set ($n = 1000$ candidates).

\subsubsection{Evaluation Metrics}
\label{app:eval-metrics}

We evaluate generated and optimized molecules using a comprehensive set of metrics:

\paragraph{Basic set metrics.}
\begin{itemize}
    \item \textbf{Validity:} Fraction of generated molecules that are parseable by RDKit.
    \item \textbf{Uniqueness:} Fraction of unique canonical SMILES among valid generated molecules.
    \item \textbf{Novelty:} Fraction of unique valid molecules not present in the training set.
\end{itemize}

\paragraph{Hypervolume metrics.}
\begin{itemize}
    \item \textbf{Hypervolume (HV):} Volume of objective space dominated by the Pareto front, bounded by a reference point. Computed using the pymoo library \citep{blank2020pymoo}.
    \item \textbf{Hypervolume improvement (HVI):} $\text{HVI} = \text{HV}(\text{baseline} \cup \text{optimized}) - \text{HV}(\text{baseline})$.
    \item \textbf{Reference point:} Automatically selected with 10\% margin beyond worst observed values: $(0.033, 7.90)$ for (QED, SA).
    \item \textbf{Bootstrap confidence intervals:} 95\% CIs computed over 1,000 bootstrap resamples of the Pareto front.
\end{itemize}

\paragraph{Surrogate fidelity.}
\begin{itemize}
    \item \textbf{MSE:} Mean squared error between surrogate predictions and oracle (RDKit-computed) values on optimized molecules.
    \item \textbf{$R^2$:} Coefficient of determination. Negative values indicate the surrogate predicts worse than a constant baseline, signaling over-optimization beyond the surrogate's reliable regime.
\end{itemize}

\paragraph{Distribution metrics.}
\begin{itemize}
    \item \textbf{Fr{\'e}chet distance (FD-FP):} Fr{\'e}chet distance computed on Morgan fingerprint embeddings (radius 2, 2048 bits) projected to 256 dimensions via random Gaussian projection. Measures distributional shift from the training set.
    \item \textbf{Descriptor KL divergences:} KL divergence $D_{\text{KL}}(P_{\text{gen}} \| P_{\text{train}})$ computed for seven molecular descriptors using 50-bin histograms:
    \begin{itemize}
        \item \textbf{MolWt:} Molecular weight (Daltons)
        \item \textbf{MolLogP:} Crippen octanol-water partition coefficient
        \item \textbf{HBD:} Number of hydrogen bond donors
        \item \textbf{HBA:} Number of hydrogen bond acceptors
        \item \textbf{TPSA:} Topological polar surface area (\AA$^2$)
        \item \textbf{RingCount:} Number of rings
        \item \textbf{RotBonds:} Number of rotatable bonds
    \end{itemize}
    \item \textbf{Average descriptor KL:} Mean KL divergence across all seven descriptors.
\end{itemize}

\paragraph{Scaffold metrics.}
\begin{itemize}
    \item \textbf{Scaffold diversity:} Fraction of unique Bemis-Murcko scaffolds \citep{bemis1996} among valid molecules. Values near 1.0 indicate high structural diversity; low values indicate scaffold collapse.
\end{itemize}

\subsection{Gamma Sweep Experimental Details}
\label{app:gamma_sweep}

This section provides additional details on the guidance strength ($\gamma$) characterization experiment described in Section~\ref{sec:pareto-advancement}.

\subsubsection{Two-Phase Search Strategy}

We employ a two-phase search to efficiently identify the optimal guidance strength without exhaustive enumeration.

\paragraph{Phase 1: Coarse Sweep.}
The first phase evaluates gamma values spanning five orders of magnitude on a logarithmic scale:
\begin{equation}
\gamma \in \{0.001, 0.01, 0.1, 1.0, 10, 100, 1000\}
\end{equation}
Each configuration is evaluated with three random seeds (42, 123, 456) to quantify variance. This coarse sweep identifies $\gamma = 100$ as the best-performing region.

\paragraph{Phase 2: Bayesian Optimization.}
The second phase refines the search around the best coarse region using tree-structured Parzen estimators (TPE). The search range is set to $\pm 3\times$ the best coarse gamma:
\begin{equation}
\gamma \in [33.3, 300.0]
\end{equation}
We run 30 Optuna trials, each evaluated with a single seed. The objective is to maximize hypervolume improvement (HVI). This phase identifies $\gamma^* = 92.4$ as optimal, achieving HVI $= 0.405$.

\subsubsection{Optimization Hyperparameters}

We use the same hyperparameters outlined in \ref{app:budgeted-hpo} with the exception of varying the guidance $\gamma$.

\subsubsection{Full Results}

Table~\ref{tab:gamma_full} presents the complete results across all evaluated gamma values, including hypervolume improvement, surrogate model fidelity, and distributional metrics. Table~\ref{tab:gamma_kl_full} provides the per-descriptor KL divergences measuring distributional shift from the training data.

\begin{table*}[ht!]
\centering
\small
\setlength{\tabcolsep}{2.2pt}
\caption{Complete gamma sweep results. HVI: hypervolume improvement. CI: mean $\pm$ std across 3 seeds. Unique/Novel: fraction among valid molecules. All configurations achieve 100\% validity except $\gamma=1000$ (unique: 13\%).}
\label{tab:gamma_full}
\begin{tabular}{lccccccccc}
\toprule
$\gamma$ & HVI$\uparrow$ & HVI (\%)$\uparrow$ & HVI CI$\uparrow$ & MSE$\downarrow$ & Unique (\%)$\uparrow$ & Novel (\%)$\uparrow$ & Scaf. Div.$\uparrow$ & FD-FP$\downarrow$ & Avg. KL$\downarrow$ \\
\midrule
$0.001$ & $0.023$ & $0.39$ & $0.023 \pm 0.016$ & $0.63$ & $100$ & $100$ & $0.962$ & $8.89$ & $0.18$ \\
$0.1$   & $0.023$ & $0.39$ & $0.023 \pm 0.016$ & $0.64$ & $100$ & $100$ & $0.963$ & $8.91$ & $0.18$ \\
$1$     & $0.023$ & $0.39$ & $0.023 \pm 0.016$ & $0.65$ & $100$ & $100$ & $0.961$ & $8.92$ & $0.18$ \\
$10$    & $0.019$ & $0.33$ & $0.019 \pm 0.015$ & $0.77$ & $100$ & $100$ & $0.956$ & $9.00$ & $0.18$ \\
$\mathbf{92.4^*}$ & $\mathbf{0.405}$ & $\mathbf{6.96}$ & --- & $1.98$ & $100$ & $100$ & $0.743$ & $13.18$ & $0.47$ \\
$100$   & $0.357$ & $6.13$ & $0.357 \pm 0.077$ & $2.02$ & $100$ & $100$ & $0.702$ & $13.46$ & $0.53$ \\
$1000$  & $0.170$ & $2.91$ & $0.170 \pm 0.003$ & $4.52$ & $13$ & $100$ & $0.000$ & $40.81$ & $8.99$ \\
\bottomrule
\end{tabular}
\end{table*}

\begin{table*}[ht!]
\centering
\small
\setlength{\tabcolsep}{5pt}
\caption{Individual descriptor KL divergences $D_{\text{KL}}(P_{\text{gen}} \| P_{\text{train}})$$\downarrow$ across guidance strengths.}
\label{tab:gamma_kl_full}
\begin{tabular}{lccccccc}
\toprule
$\gamma$ & MolWt & MolLogP & HBD & HBA & TPSA & RingCount & RotBonds \\
\midrule
$0.001$ & $0.22$ & $0.13$ & $0.02$ & $0.16$ & $0.14$ & $0.12$ & $0.48$ \\
$0.1$ & $0.22$ & $0.12$ & $0.02$ & $0.16$ & $0.14$ & $0.12$ & $0.46$ \\
$1$ & $0.22$ & $0.13$ & $0.02$ & $0.15$ & $0.15$ & $0.12$ & $0.46$ \\
$10$ & $0.24$ & $0.13$ & $0.02$ & $0.17$ & $0.14$ & $0.11$ & $0.42$ \\
$\mathbf{92.4^*}$ & $0.68$ & $0.22$ & $0.03$ & $0.76$ & $0.67$ & $0.27$ & $0.68$ \\
$100$ & $0.76$ & $0.23$ & $0.05$ & $0.86$ & $0.77$ & $0.32$ & $0.71$ \\
$1000$ & $16.91$ & $15.54$ & $0.67$ & $3.36$ & $5.04$ & $5.19$ & $16.20$ \\
\bottomrule
\end{tabular}
\end{table*}

\subsubsection{Interpretation of Results}

The gamma sweep reveals three distinct operating regimes:

\paragraph{Conservative regime ($\gamma \leq 10$).}
Guidance is too weak to overcome the flow prior. Molecules remain near their original positions in latent space, resulting in minimal property changes (HVI $< 0.4\%$) but excellent scaffold preservation (diversity $> 0.95$) and low distributional shift (Fr\'echet $< 9.1$).

\paragraph{Optimal regime ($\gamma \approx 50$--$150$).}
Guidance is strong enough to push molecules toward improved property regions while the surrogate model remains predictive. The optimal $\gamma^* = 92.4$ achieves maximum HVI (6.96\%) with acceptable trade-offs: scaffold diversity remains at 74\%, and Fr\'echet distance increases moderately to 13.2.

\paragraph{Over-optimization regime ($\gamma > 200$).}
Guidance dominates the flow prior, causing optimization trajectories to escape the surrogate's reliable predictive regime. Symptoms include: (1) decreasing HVI despite stronger guidance, (2) collapsing scaffold diversity (0\% at $\gamma = 1000$), (3) severe surrogate failure ($R^2 = -40.2$), and (4) large distributional shift (Fr\'echet $= 40.8$).

The non-monotonic relationship between $\gamma$ and HVI---rising until $\gamma \approx 100$ then declining---is the signature of surrogate exploitation. At extreme $\gamma$, the optimizer finds molecules that score well according to the surrogate but poorly according to the oracle, having left the manifold where the surrogate was trained.

\newpage

\subsubsection{Hyperparameter Summary}

Table~\ref{tab:hyperparams} summarizes key hyperparameters.

\begin{table}[ht!]
\centering
\small
\caption{Hyperparameters for ZINC optimization experiments.}
\label{tab:hyperparams}
\begin{tabular}{ll}
\toprule
\textbf{Parameter} & \textbf{Value} \\
\midrule
\multicolumn{2}{l}{\textit{Data}} \\
Dataset & ZINC250K \\
Representation & SELFIES \\
Max sequence length & 128 \\
Vocabulary size & 111 \\
Train/test split & 81/9/10 \\
\midrule
\multicolumn{2}{l}{\textit{VAE (Pretrain)}} \\
Encoder/decoder layers & 6 \\
Hidden dimension ($d_{\text{model}}$) & 128 \\
Attention heads & 8 \\
Feedforward dimension & 1024 \\
Latent dimension & 128 \\
Codebook size ($K$) & 8 \\
Dropout & 0.1 \\
Pretrain epochs & 150 \\
Batch size & 256 \\
Learning rate & $10^{-4}$ \\
KL weight ($\beta_{\max}$) & 0.1 \\
KL warmup fraction & 0.35 \\
\midrule
\multicolumn{2}{l}{\textit{VAE (Finetune)}} \\
Finetune epochs & 20 \\
Batch size & 1024 \\
Learning rate & $10^{-3}$ \\
Property weight & 1.0 \\
\midrule
\multicolumn{2}{l}{\textit{Surrogate}} \\
Hidden dimension & 1024 \\
Aggregation & Mean pooling \\
Dropout & 0.1 \\
\midrule
\multicolumn{2}{l}{\textit{Flow model}} \\
Transformer layers & 10 \\
Hidden dimension ($d_{\text{model}}$) & 256 \\
Attention heads & 8 \\
Feedforward dimension & 512 \\
Time embedding dimension & 128 \\
Dropout & 0.1 \\
Training epochs & 100 \\
Batch size & 1024 \\
Learning rate & $2 \times 10^{-4}$ \\
\midrule
\multicolumn{2}{l}{\textit{Optimization}} \\
Candidates ($n$) & 1000 \\
Noise scale ($\sigma$) & 0.80 \\
Integration start ($t_{\text{start}}$) & 0.89 \\
Integration steps & 12 \\
\midrule
\multicolumn{2}{l}{\textit{Evaluation}} \\
Reference point margin & 0.1 \\
Bootstrap samples & 1000 \\
Confidence level & 0.95 \\
\bottomrule
\end{tabular}
\end{table}

\newpage

\subsection{Molecular generation and ablations}
\label{app:ablations}

Additional ablation results and qualitative analyses of MoltenFlow’s molecular generation behavior, complementing the optimization-focused experiments in the main text.

% ------------------- FIGURE (A2) -------------------
% Replace filenames with whatever you copied into your figures/ directory.
% Note: each panel is "fit on real" within its own model run; axes are not directly comparable.
\begin{figure}[ht!]
\centering
\begin{subfigure}[t]{0.49\linewidth}
  \centering
  \includegraphics[width=\linewidth]{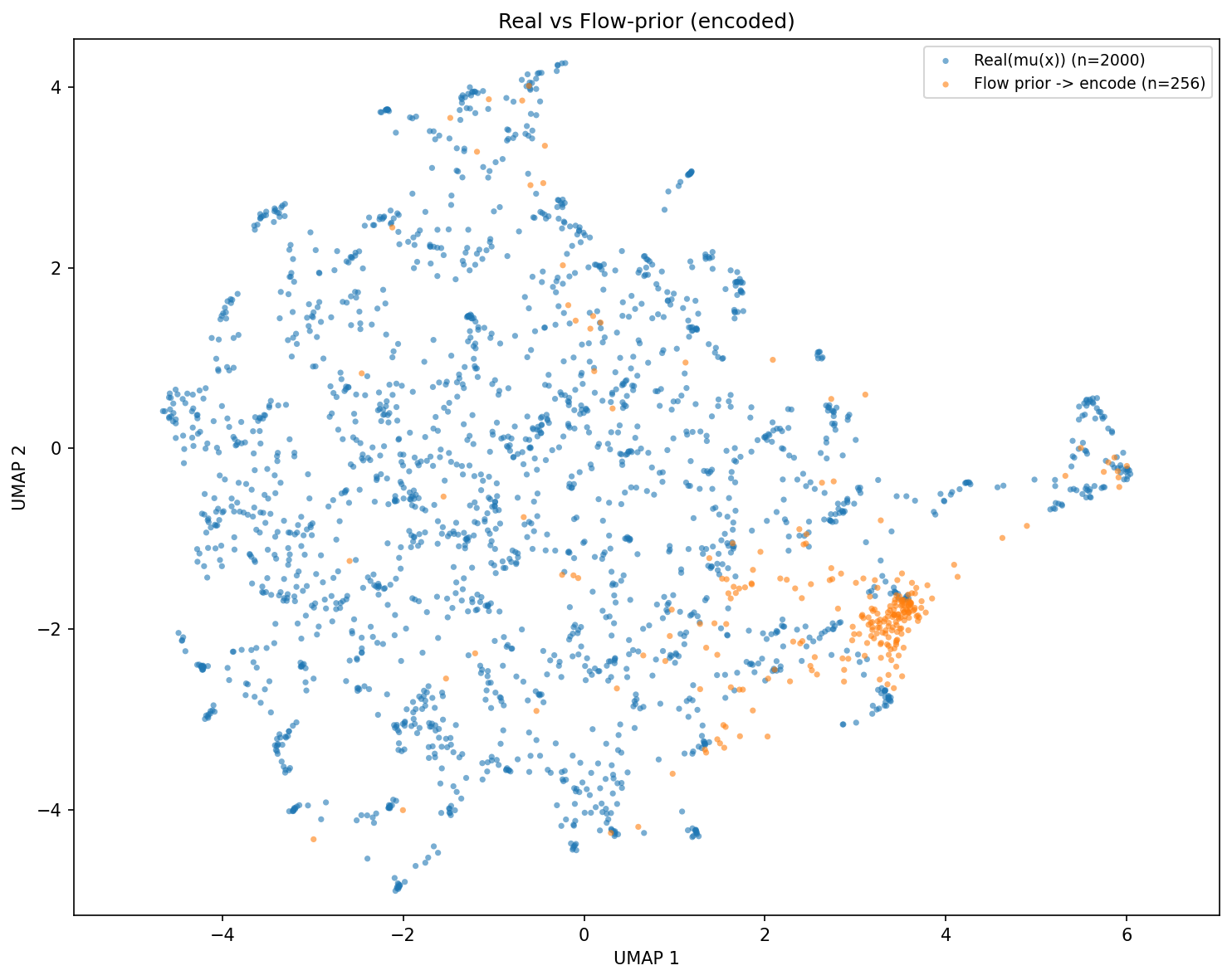}
  \caption{SMILES model: real vs flow prior (re-encoded).}
\end{subfigure}\hfill
\begin{subfigure}[t]{0.49\linewidth}
  \centering
  \includegraphics[width=\linewidth]{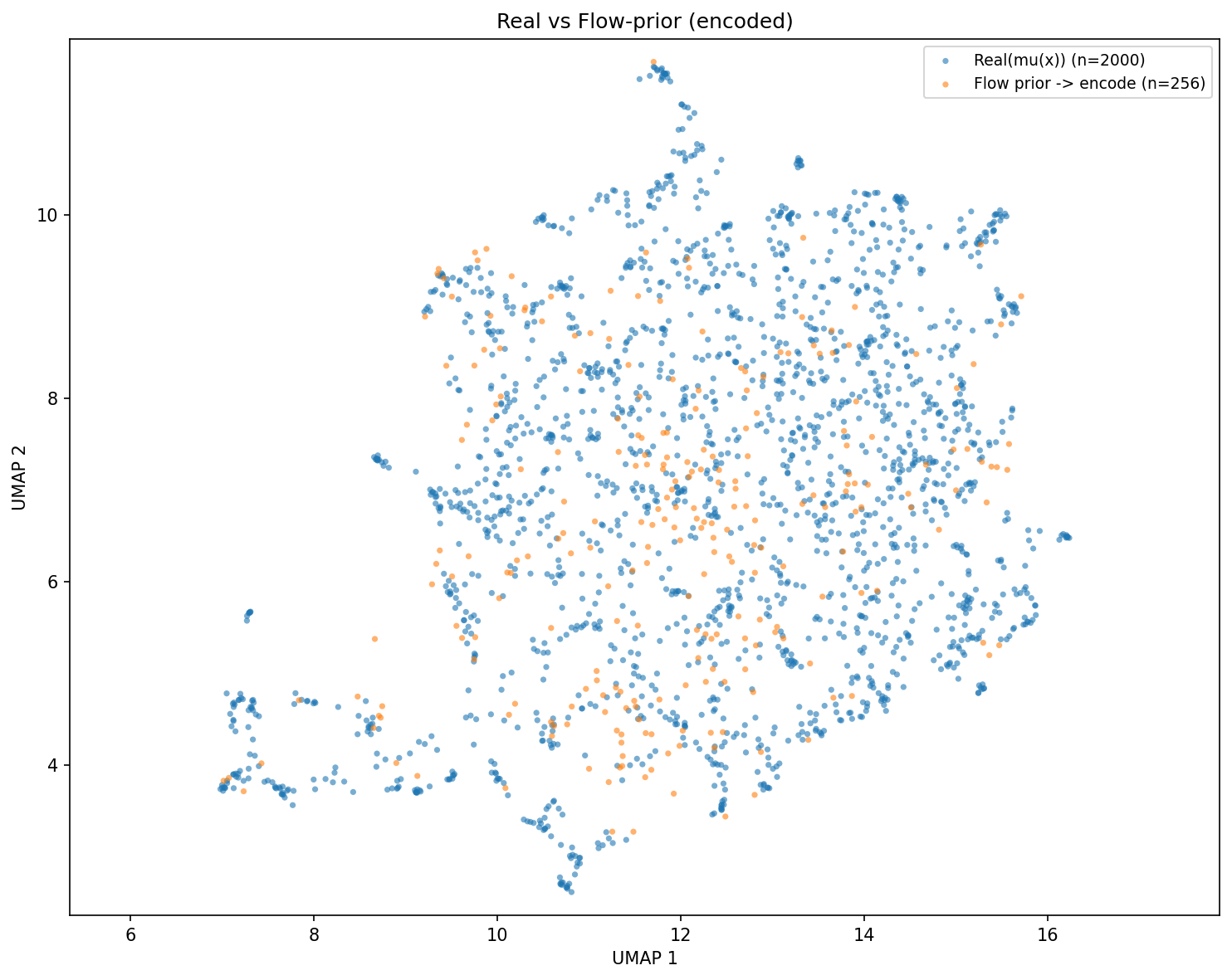}
  \caption{SELFIES model: real vs flow prior (re-encoded).}
\end{subfigure}
\caption{UMAP of posterior-mean latent embeddings $\mu_\phi(x)$ for real molecules (fit set) and generated molecules after re-encoding.
Each UMAP is fit on the real embeddings for the corresponding model run.}
\label{fig:umap_a2_smiles_vs_selfies}
\end{figure}

\end{document}